\documentclass[twoside,11pt]{article}

\usepackage{jmlr2e_mod}
\usepackage{amsmath}
\usepackage{color}

\usepackage{subcaption}
\usepackage{booktabs}

\usepackage{etoc}

\newcommand{\ErrorGen}{\error_{\text{gen}}}
\newcommand{\ErrorOpt}{\error_{\text{opt}}}
\newcommand{\ErrorSta}{\error_{\text{stab}}}

\newcommand{\matsnorm}[2]{|\!|\!| #1 | \! | \!|_{{#2}}}
\newcommand{\vecnorm}[2]{\left\| #1\right\|_{#2}}

\newcommand{\Exs}{\ensuremath{{\mathbb{E}}}}
\newcommand{\Prob}{\ensuremath{{\mathbb{P}}}}
\newcommand{\error}{\ensuremath{\mathcal{E}}}

\newcommand{\NORMAL}{\ensuremath{\mathcal{N}}}

\newcommand{\defeq}{\ensuremath{\equiv}}

\newcommand{\Zspace}{\ensuremath{\mathcal{Z}}}

\newcommand{\Thetaspace}{\ensuremath{\Omega}}
\newcommand{\samples}{\ensuremath{S}}

\newcommand{\thetabasic}{\ensuremath{\theta}}
\newcommand{\thetastar}{\ensuremath{{\theta^*}}}
\newcommand{\thetahat}{\ensuremath{{\hat{\theta}}}}
\newcommand{\thetatil}{\ensuremath{{\tilde{\theta}}}}
\newcommand{\thetahatS}{\ensuremath{{\hat{\theta}_\samples}}}

\newcommand{\thetazero}{\ensuremath{{\theta_0}}}

\newcommand{\loss}{\ensuremath{l}}
\newcommand{\lossf}{\ensuremath{f}}
\newcommand{\totalrisk}{\ensuremath{R}}

\newcommand{\yvec}{\ensuremath{Y}}
\newcommand{\Xmat}{\ensuremath{\mathbf{X}}}

\newcommand{\Ind}{\ensuremath{\mathbb{I}}}

\newcommand{\iters}{\ensuremath{T}}
\newcommand{\obs}{\ensuremath{n}}
\newcommand{\dims}{\ensuremath{d}}

\newcommand{\real}{\ensuremath{\mathbb{R}}}

\newcommand{\brackets}[1]{\left[ #1 \right]}
\newcommand{\parenth}[1]{\left( #1 \right)}
\newcommand{\braces}[1]{\left\{ #1 \right \}}
\newcommand{\abss}[1]{\left| #1 \right |}
\newcommand{\angles}[1]{\left\langle #1 \right \rangle}

\newcommand{\distribution}{\ensuremath{P}}
\newcommand{\distributions}{\ensuremath{\mathcal{P}}}
\newcommand{\losses}{\ensuremath{\mathcal{L}}}
\newcommand{\lossesconvex}{\ensuremath{\losses_{\text{c}}}}
\newcommand{\lossesstronglyconvex}{\ensuremath{\losses_{\text{sc}}}}

\newcommand{\CCerm}{\ensuremath{C_1}}
\newcommand{\CCopt}{\ensuremath{C_2}}
\newcommand{\CCermstrong}{\ensuremath{C_3}}

\newcommand{\CC}{\ensuremath{C}}
\newcommand{\temperature}{\ensuremath{\tau}}

\ShortHeadings{Stability and Convergence Trade-off of Iterative Optimization Algorithms}{Yuansi Chen, Chi Jin and Bin Yu}
\firstpageno{1}

\begin{document}\thispagestyle{empty}

\etocdepthtag.toc{mtchapter}
\etocsettagdepth{mtchapter}{subsection}
\etocsettagdepth{mtappendix}{none}

\title{Stability and Convergence Trade-off of Iterative Optimization Algorithms}

\author{\name Yuansi Chen \email yuansi.chen@berkeley.edu \\
       \addr Department of Statistics\\
       University of California\\
       Berkeley, CA 94720-1776, USA
       \AND
       \name Chi Jin \email chijin@berkeley.edu \\
       \addr Department of Electrical Engineering and Computer Sciences\\
       University of California\\
       Berkeley, CA 94720-1776, USA
       \AND
       \name Bin Yu \email binyu@berkeley.edu \\
       \addr Department of Statistics \& Department of Electrical Engineering and Computer Sciences\\
       University of California\\
       Berkeley, CA 94720-1776, USA}

\editor{......}

\maketitle

\begin{abstract}%
The overall performance or expected excess risk of an iterative machine learning algorithm can be decomposed into training error and generalization error. While the former is controlled by its convergence analysis, the latter can be tightly handled by algorithmic stability~\citep{bousquet2002stability}. The machine learning community has a rich history investigating convergence and stability separately. However, the question about the  trade-off between these two quantities remains open.

In this paper, we show that for any iterative algorithm at any iteration, the overall performance is lower bounded by the minimax statistical error over an appropriately chosen loss function class.
This implies an important trade-off between convergence and stability of the algorithm -- a faster converging algorithm has to be less stable, and vice versa. As a direct consequence of this fundamental tradeoff, new convergence lower bounds can be derived for classes of algorithms constrained with different stability bounds. In particular, when the loss function is convex (or strongly convex) and smooth, we discuss the stability upper bounds of gradient descent (GD) and stochastic gradient descent and their variants with decreasing step sizes. For Nesterov's accelerated gradient descent (NAG) and heavy ball method (HB), we provide stability upper bounds for the quadratic loss function. Applying existing stability upper bounds for the gradient methods in our trade-off framework, we obtain lower bounds matching the well-established convergence upper bounds up to constants for these algorithms and conjecture similar lower bounds for NAG and HB. Finally, we numerically demonstrate the tightness of our stability bounds in terms of exponents in the rate and also illustrate via a simulated logistic regression problem that our stability bounds reflect the generalization errors better than the simple uniform convergence bounds for GD and NAG.
\end{abstract}


\section{Introduction} 
\label{sec:introduction}
For different supervised learning algorithms ranging from classical linear regression, logistic regression, boosting, to modern large-scale deep networks, the overall performance or expected excess risk can always be decomposed into two parts:
the empirical error (or the training error) and the generalization error (characterizing the discrepancy between the test error and the training error). A central theme in machine learning is to find an appropriate balance between empirical error and generalization error, because improperly emphasizing one over the other typically results in either overfitting or underfitting. Specifically, in the context of supervised learning models trained by iterative optimization algorithms, the empirical error at each iteration is commonly controlled by convergence rate analysis, and the generalization error can be handled by algorithmic stability analysis~\citep{devroye1979distribution, bousquet2002stability}.

Convergence rate of an algorithm portrays how fast the optimization error decreases as the number of iterations grows.
Recent years have witnessed a rapid advance on convergence rates analysis of specific optimization methods for a particular class of loss functions that they are optimizing over. In fact, such analysis has been carried out for many gradient methods, including gradient descent (GD), Nesterov accelerated gradient descent (NAG), stochastic gradient descent (SGD), stochastic gradient Langevin dynamics (SGLD) for convex, strongly convex, or even nonconvex functions (see e.g.~\citet{boyd2004convex,bubeck2015convex,nesterov2013introductory,jin2017escape,raginsky2017non}).
However, until the optimization error and generalization error of these algorithms are analyzed together, it is not clear whether the fastest converging optimization algorithm is the best for learning.

On the other hand, algorithmic stability~\citep{devroye1979distribution, bousquet2002stability} in learning problems has been introduced as an alternative way to control generalization error instead of uniform convergence results such as classical VC-theory~\citep{vapnik1994measuring} and Rademacher complexity~\citep{bartlett2003rademacher}. The stability concept has an intuitive appeal: an algorithm is stable if it is robust to small perturbations in the composition of the learning data set. Recently it has been shown that algorithmic stability is well suited for controlling generalization error of stochastic gradient methods~\citep{hardt2015train}, as well as stochastic gradient Langevin dynamics algorithm~\citep{mou2017generalization}.

While most previous papers study convergence rate and the algorithmic stability of an optimization algorithm separately, a natural question arises: What is the relationship or trade-off between the convergence rate and the algorithmic stability of an iterative algorithm? Is it possible to design an algorithm that converges the fastest and at the same time most stable? If not, is there any fundamental limit on the trade-off between the two quantities so that a fast algorithm has to be unstable?

This paper shows that there is a fundamental limit on the trade-off. That is, for any iterative algorithms, at any time step, the sum of optimization error and stability is lower bounded by the minimax statistical error over a given loss function class. Therefore, a fast converging algorithm can not be too stable, and a stable algorithm can not converge too fast. This framework therefore provides a new criterion for comparing optimization algorithms by considering jointly convergence rate and algorithm stability. As a consequence, our framework can be immediately applied to provide a new class of convergence lower bounds for algorithms with different stability rates.

In particular, we focus on two settings where the loss functions are either convex smooth or strongly convex smooth.
In the first setting, we discuss the stability upper bounds of gradient descent (GD), stochastic gradient descent (SGD) and their variants with decreasing step sizes. New stability upper bounds are provided for Nesterov's accelerated gradient descent (NAG) and the heavy ball method (HB) under quadratic loss, and we conjecture these upper bounds still hold for the general convex smooth losses. Applying the stability upper bounds for GD and SGD in our trade-off framework, we obtain the convergence lower bounds for them that match the known convergence upper bounds up to constants. Considering jointly convergence rate and algorithm stability for NAG and GD, the trade-off shows that NAG must be less stable than GD even though it converges faster than GD.
In the second setting where the loss functions are strongly convex and smooth, we also provide stability upper bound and deduce the convergence lower bound results for GD and NAG via our trade-off framework. Finally, simulations are conducted to show that the stability bounds established have the correct rates as a function of $\obs$ and iteration $\iters$. These bounds are demonstrated to be particularly useful in large scale learning settings for understanding the overall performance of an algorithm than the classical uniform convergence bounds because the stability bounds capture better generalization errors at early iterations of these algorithms.

\subsection{Related work} 
\label{sub:related_work}
\paragraph{Algorithmic stability} The first quantitative results that focus on generalization error via algorithmic stability date back to~\citep{rogers1978finite, devroye1979distribution}. This line of research was further developed by~\citet{bousquet2002stability} to provide guarantees for general supervised learning algorithms and insights for the practice of regularized algorithms. It remains unclear, however, what is the algorithmic stability of general iterative optimization algorithms. Recently, to show the effectiveness of commonly used optimization algorithms in many large-scale learning problems, algorithmic stability has been established for stochastic gradient methods \citep{hardt2015train}, stochastic gradient Langevin dynamics \citep{mou2017generalization}, as well as for any algorithm in situations where global minima are approximately achieved \citep{charles2017stability}.

\paragraph{Lower bounds on convergence rate} Given the importance of efficient optimization methods, many papers have been devoted to understanding the fundamental computational limits of convex optimization. Those lower bounds typically focus on a specific class of algorithms. A classical line of research has been focused on first-order algorithms where only first-order information (i.e. gradients) can be queried through oracle model; see the book~\citep{boyd2004convex}, the monograph~\citep{bubeck2015convex} and references therein for further details.
For convex functions, the first lower bound argument given in~\citep{nemirovsky1982problem} applies to first-order algorithms whose current iterate lies in the linear span of previous gradients. It has been later extended to any deterministic, then stochastic first-order algorithm~\citep{agarwal2014lower,woodworth2016tight}.

\subsection{Organization of the paper}
The rest of the paper is organized as follows: In Section~\ref{sec: Background and preliminaries}, we set up the necessary backgrounds on the classical excess risk decomposition and introduce the optimization error (or computational bias) and generalization error trade-off. In Section~\ref{sec: Trade-off between stability and convergence rate}, we provide the main theorem on the trade-off between convergence rate (as an upper bound on optimization error) and algorithmic stability (as an upper bound on generalization error). In Section~\ref{sec: Stability of first order optimization algorithms}, we establish uniform stability bounds for several gradient methods and show that our main theorem applies to these algorithms to obtain their convergence lower bounds. In Section~\ref{sec:simulations}, we first provide simulation results validating the correct rates as a function of sample size $\obs$ and iteration number $\iters$ of the stability bounds we established, and then illustrate via a simulated logistic regression problem that our stability bounds reflect the generalization errors better than the simple uniform convergence bounds for GD and NAG.


\section{Preliminaries}
\label{sec: Background and preliminaries}
In this section, we set up the necessary backgrounds on excess risk decomposition and convex optimization. Using classical excess risk decomposition, we introduce the expected optimization error and generalization error trade-off which are crucial to state our main result in the next section.
\subsection{Excess risk decomposition}
Throughout this paper, we consider the standard setting of supervised learning. Suppose that we are given $\obs$ samples $\samples = \parenth{z_1,...,z_\obs}$, each lying in some space $\Zspace$ and drawn i.i.d. according to a distribution $\distribution \in \distributions$. The standard decision-theoretic approach is to estimate a parameter $\thetabasic\in \real^\dims$ by minimizing a loss function of the form $\loss(\thetabasic; z)$, which measures the fit between the model indexed by the parameter $\thetabasic \in \Thetaspace\subset \real^\dims$ and the sample $z \in \Zspace$.

Given the collection $\samples$ of $\obs$ samples and a loss function $l$, the principle of empirical risk minimization is based on the objective function
\begin{align*}
\totalrisk_S\parenth{\thetabasic} \defeq \frac{1}{\obs} \sum_{i=1}^\obs \loss\parenth{\thetabasic; z_i}.
\end{align*}
This empirical risk above serves as a sample-average proxy for the population risk
\begin{align*}
\totalrisk(\thetabasic) \defeq \Exs_{z \sim \distribution}\brackets{\loss\parenth{\thetabasic; z}}.
\end{align*}

We denote by $\thetahat$ an estimator computed from sample $\samples$. The statistical question is how to bound the excess risk, measured in terms of the difference between the population risk and the minimal risk over the entire parameter space $\Thetaspace$,
\begin{align*}
\delta \totalrisk(\thetahat) \defeq \totalrisk(\thetahat) - \inf_{\thetabasic \in \Thetaspace} \totalrisk(\thetabasic).
\end{align*}
In most of our analysis, $\thetahat$ is the output of an optimization algorithm at a particular iteration $\iters$ based on sample $\samples$. We further denote $\thetatil$ an empirical risk minimizer. Note that $\thetahat$ and $\thetatil$ are in general not the same estimator.

For simplicity, we assume that there exists some $\thetazero \in \Thetaspace$ such that $\totalrisk(\thetazero) = \inf_{\thetabasic \in \Thetaspace} \totalrisk(\thetabasic)$.\footnote{If the infimum is not achieved within $\Thetaspace$ (for example $\Thetaspace$ is an open set), we can choose some $\thetazero$ where this equality holds up to some arbitrarily small error.}

Controlling the excess risk of the estimator $\thetahat$ is usually done by decomposing it into three terms as follows:
\begin{align*}
\delta \totalrisk(\thetahat) = \underbrace{\totalrisk(\thetahat) - \totalrisk_\samples(\thetahat)}_{T_1} + \underbrace{\totalrisk_\samples(\thetahat) - \totalrisk_\samples(\thetazero)}_{T_2} + \underbrace{\totalrisk_\samples(\thetazero) - \totalrisk(\thetazero)}_{T_3}.
\end{align*}
Term $T_1$ is the generalization error of the model $\thetahat$. Term $T_2$ is the empirical risk difference between the model $\thetahat$ and the population risk minimizer $\thetazero$. Term $T_3$ is the generalization error of $\thetazero$.

Taking expectation on the previous decomposition and noticing that $\mathbb{E}_\samples\brackets{T_3} = 0$, we obtain first a decomposition of the expected excess risk and then an upper bound:
\begin{align*}
\Exs_\samples[\delta \totalrisk(\thetahat)] &= \Exs_\samples[\totalrisk(\thetahat) - \totalrisk_\samples(\thetahat)] + \Exs_\samples[\totalrisk_\samples(\thetahat) - \totalrisk_\samples(\thetazero)] + 0\\
&= \underbrace{\Exs_\samples[\totalrisk(\thetahat) - \totalrisk_\samples(\thetahat)]}_{\ErrorGen} + \underbrace{\Exs_\samples[\totalrisk_\samples(\thetahat) - \totalrisk_\samples(\thetatil)]}_{\ErrorOpt} + \underbrace{\Exs_\samples[\totalrisk_\samples(\thetatil) - \totalrisk_\samples(\thetazero)]}_{\leq 0}\\
& \leq \underbrace{\Exs_\samples[\totalrisk(\thetahat) - \totalrisk_\samples(\thetahat)]}_{\ErrorGen} + \underbrace{\Exs_\samples[\totalrisk_\samples(\thetahat) - \totalrisk_\samples(\thetatil)]}_{\ErrorOpt}.
\end{align*}
The last inequality follows from the fact that $\thetatil$ is the empirical risk minimizer.
Hence, the expected excess risk is upper bounded by the sum of expected generalization error and the \textit{expected optimization error} or \textit{computational bias}  $\Exs_\samples[\totalrisk_\samples(\thetahat) - \totalrisk_\samples(\thetatil)]$. We formally define these two quantities indexed by the estimator $\hat{\theta}$, loss function $l$, data distribution $\distribution$ and sample size $\obs$ to be
\begin{align*}
    \ErrorGen(\hat{\theta}, l, \distribution, \obs) \equiv \Exs_{\samples\sim \distribution^\obs}\brackets{\totalrisk(\thetahat) - \totalrisk_\samples(\thetahat)},
\end{align*}
and
\begin{align*}
     \ErrorOpt(\hat{\theta}, l, \distribution, \obs) \equiv \Exs_{\samples\sim \distribution^\obs}\brackets{\totalrisk_\samples(\thetahat) - \totalrisk_\samples(\thetatil)}.
\end{align*}

Making the optimization error appear in the decomposition is useful for analyzing optimization algorithms in an iterative manner. As noted in~\citet{bousquet2008tradeoffs}, introducing optimization error allows to analyze algorithms doing approximate optimization. However, our framework is different to that introduced by~\cite{bousquet2008tradeoffs}. We control the generalization error via iteration-dependent algorithmic stability instead of directly invoking uniform convergence results. As we are going to show, for most iterative optimization algorithms, upper bounding the generalization error by a simple uniform convergence is often loose and algorithmic stability can serve as a tighter bound.

\subsection{Algorithmic Stability}
Many forms of algorithmic stability have been introduced to characterize generalization error~\citep{bousquet2002stability,kutin2002almost}. For the purpose of this paper, we are only interested in the \textit{uniform stability} notion introduced by~\citet{bousquet2002stability}.
\begin{definition}
An algorithm, which outputs a model $\thetahatS$ for sample $\samples$, is \textbf{$\epsilon$-uniform stable} if for all $k \in \{ 1,..., \obs \}$, for all data sample pair $\samples = (z_1, ..., z_k, ..., z_\obs)$ and $\samples' = (z_1, ..., z_k', ..., z_\obs)$, each $z_i$ or $z_k'$ is i.i.d sampled from $\distribution$, we have
\begin{align}
    \label{eq: stability_def}
    \sup_{z \in \Zspace} \left| l(\thetahatS; z) - l(\thetahat_{S'}; z) \right| \leq \epsilon.
\end{align}
\end{definition}
As we did for the generalization error, we use $\ErrorSta(\hat{\theta}, l, P, \obs)$ to denote the uniform stability of an algorithm $\hat{\theta}$.

A stable algorithm has the property that removing one element in its learning data set does not change much of its outcome. Such a data perturbation scheme is closely related to Jackknife in statistics~\citep{efron1982jackknife}. One can further show that uniform stability implies expected generalization~\citep{bousquet2002stability} . For completeness, we reformulate this property in the following lemma.
\begin{lemma}
\label{prop_stability}
An algorithm, which outputs a model $\thetahatS$ for sample $\samples$, is $\epsilon$-uniformly stable, then its expected generalization error is bounded as follows,
\begin{align*}
\abss{\Exs_\samples\brackets{\totalrisk(\thetahatS) - \totalrisk_\samples(\thetahatS)} } \leq \epsilon.
\end{align*}
\end{lemma}
Lemma~\ref{prop_stability} implies that $\ErrorGen(\hat{\theta}, l, \distribution, \obs) \leq \ErrorSta(\hat{\theta}, l, \distribution, \obs)$. The proof provided by~\cite{bousquet2002stability} relies on a symmetrization argument and makes use of the i.i.d assumptions of samples in $\samples$. Combining the expected excess risk decomposition in previous section, we conclude that the sum of uniform stability and expected optimization error (or computational bias) constitutes an upper bound for the expected excess risk,
\begin{align}
\label{eq: risk}
\Exs_{\samples\sim \distribution^\obs}[\delta \totalrisk(\thetahatS)] \leq \ErrorSta(\hat{\theta}, l, \distribution, \obs) + \ErrorOpt(\hat{\theta}, l, \distribution, \obs).
\end{align}
Note that the result is stated for a fixed loss function $l$ and a fixed data distribution $\distribution$.
Equation~\eqref{eq: risk} is a key inequality for our analysis. Not only it provides a way to upper bound the expected excess risk without uniform convergence results, but also it makes the connection between the statistical excess risk and the optimization convergence rate (or computational bias). This can also be seen as reminiscent of the bias-variance trade-off of an algorithm in a computational sense since stability serves as a computational variability term and optimization error as a computational bias term.

\subsection{Convex optimization settings}
Throughout the paper, we focus on two types of loss functions: The first type of loss function $\loss(\cdot, z)$ is $\alpha$-strongly convex and $\beta$-smooth for every $z$; The second type of loss function $\loss(\cdot, z)$ is convex and $\beta$-smooth for every $z$. We also make use of the $L$-Lipschitz condition. We provide their definitions here. More technical details about convex optimization and relevant results are deferred to Appendix~\ref{sec:appendix_stability_bounds_for_convex_smooth_functions}.
\begin{definition}
A function  $f$ is $L$-\textit{Lipschitz} if for all $u, v \in \Thetaspace$, we have
\begin{align*}
\abss{f(u) - f(v)} \leq L \vecnorm{u - v}{2}.
\end{align*}
\end{definition}

\begin{definition}
A function  continuously differentiable $f$ is $\beta$-\textit{smooth} if for all $u, v \in \Thetaspace$, we have
\begin{align*}
\vecnorm{\nabla f(u) - \nabla f(v)}{2} \leq \beta \vecnorm{ u - v }{2}.
\end{align*}
\end{definition}

\begin{definition}
A function $f$ is \textit{convex} if for all $u, v \in \Thetaspace$,
we have
\begin{align*}
f(u) \geq f(v) + \angles{\nabla f(v), u - v}.
\end{align*}
\end{definition}

\begin{definition}
A function $f$ is $\alpha$-\textit{strongly convex} if for all $u, v \in \Thetaspace$,
we have
\begin{align*}
f(u) \geq f(v) + \angles{\nabla f(v), u - v} + \frac{\alpha}{2} \vecnorm{u-v}{2}^2.
\end{align*}
\end{definition}

\section{Trade-off between stability and convergence rate}
\label{sec: Trade-off between stability and convergence rate}
In this section, we introduce the trade-off between stability and convergence rate via excess risk decomposition under two settings of loss functions mentioned in the previous section: the convex smooth setting and the strongly convex smooth setting. We show that for any iterative algorithm, at any time step, the sum of optimization error and stability is lower bounded by the minimax statistical error over a given loss function class. Thus algorithms sharing the same stability upper bound can be grouped to obtain convergence rate lower bounds. This provides a new class of convergence lower bounds for algorithms with different stability bounds.

We are interested in distribution independent stability and convergence where we take supremum of these two quantities over distributions and losses. For a fixed iteration algorithm that outputs $\hat{\theta}$ at iteration $\iters$, we define its uniform stability and optimization error as follows,
\begin{align*}
    \ErrorSta^{\hat{\theta}}(\iters, \obs, \losses) &\equiv \sup_{l \in \losses, \distribution \in \distributions} \ErrorSta(\hat{\theta}_T, l, \distribution, \obs),\\
    \ErrorOpt^{\hat{\theta}}(\iters, \obs, \losses) &\equiv \sup_{l \in \losses, \distribution \in \distributions} \ErrorOpt(\hat{\theta}_T, \distribution, \obs).
\end{align*}
Note that in this paper, the supremum is taken over the class of all loss functions $\losses$ under either of the two settings considered (convex smooth and strongly convex smooth settings).

\subsection{Trade-off in the convex smooth setting}
Before we state the main theorem, we first define the loss function class of interest in this section. We define the class of all convex smooth loss functions as follows,
\begin{align*}
    \lossesconvex = \braces{l: \Zspace \times \Thetaspace \rightarrow \real   | l \text{ is convex, $\beta$-smooth}, \abss{\Omega} = R}.
\end{align*}
In the convex smooth setting, we have the following lower bound on the sum of stability and convergence rate.
\begin{theorem}
\label{theorem: lower_bound}
Suppose an iterative algorithm outputs $\hat{\theta}_\iters$ at iteration $\iters$ on an empirical loss built upon a loss $l \in \lossesconvex$ and an i.i.d. sample $\samples$ of size $\obs$, and it has uniform stability $\ErrorSta(\iters, \obs, \lossesconvex)$ and optimization error $\ErrorOpt(\iters, \obs, \lossesconvex)$, then there exists a universal constant $\CCerm >0 $ such that,
\begin{align*}
\ErrorSta^{\hat{\theta}}(\iters, \obs, \lossesconvex) + \ErrorOpt^{\hat{\theta}}(\iters, \obs, \lossesconvex) \geq \inf_{\tilde{\theta}} \sup_{\distribution \in \distributions}\Exs_{\samples\sim P^n}[\delta \totalrisk(\tilde{\theta})] \geq \frac{R^2 \beta }{\CCerm \sqrt{\obs}}
\end{align*}
\end{theorem}
The first inequality of Theorem~\ref{theorem: lower_bound} is a simple outcome of the empirical risk decomposition in Equation~\eqref{eq: risk}. This first inequality is not tied to the convex smooth setting and can generalize to a wide class of optimization algorithms. The second inequality is based on an adaptation of the classical~\cite{cam1986asymptotic}'s method for minimax estimation lower bound to the convex smooth loss function class. Further, if we know $\ErrorSta^{\hat{\theta}}(\iters, \obs, \lossesconvex)$ precisely, we can obtain an immediate corollary that provide convergence lower bound for stable optimization algorithms.

\begin{corollary}
\label{corollary: optimization_lower_bound}
Under conditions in Theorem~\ref{theorem: lower_bound}, if an algorithms has uniform stability
\begin{align*}
\ErrorSta^{\hat{\theta}}(\iters, \obs,  \lossesconvex) \leq \frac{s(\iters)}{\obs},
\end{align*}
with $s$ a divergent function of $\iters$, i.e.
\begin{align*}
    s(\iters) \rightarrow \infty, \text{ as }\iters \rightarrow \infty,
\end{align*}
then there exists a universal constant $\CCopt > 0$, a sample size $\obs_0$ and an iteration number $\iters_0 \geq 1$, such that for $\iters \geq \iters_0$, its convergence rate is lower bounded as follows,
\begin{align*}
\ErrorOpt^{\hat{\theta}}(\iters, \obs_0,  \lossesconvex) \geq \frac{R^4 \beta^2}{\CCopt s(\iters)}.
\end{align*}

\end{corollary}
Even though Theorem~\ref{theorem: lower_bound} is valid for any pair of $(\iters, \obs)$, Corollary~\ref{corollary: optimization_lower_bound} requires to choose a specific sample size $n_0$ in construction. However, under the assumption that the optimization algorithm has convergence rate independent of the sample size (i.e. $\ErrorOpt^{\hat{\theta}}(\iters, \obs, \lossesconvex)$ is not a function of $\obs$), we can obtain via Corollary~\ref{corollary: optimization_lower_bound} a convergence lower bound that is comparable to the lower bounds in the convex optimization literature. We remark that this assumption is satisfied for commonly-used optimization algorithms such as GD and NAG.

Theorem~\ref{theorem: lower_bound} and Corollary~\ref{corollary: optimization_lower_bound} provide the trade-off between stability and optimization convergence rate. All iterative optimization methods that are algorithmic uniform stable can not converge too fast. This motivates the idea of grouping optimization methods with their algorithmic stability. Optimization methods that share the same algorithmic stability would have the same optimization lower bound. The proof of Theorem~\ref{theorem: lower_bound} is provided in Appendix~\ref{ssec:proof_of_theorem_theorem: lower_bound} and that of Corollary~\ref{corollary: optimization_lower_bound} in Appendix~\ref{ssec:proof_of_corollary: optimization_lower_bound}.

\subsection{Trade-off in the strongly convex smooth setting}
Similar to the convex smooth setting, we define the class of all strongly convex smooth loss functions as follows,
\begin{align*}
    \lossesstronglyconvex = \braces{l: \Zspace \times \Thetaspace \rightarrow \real   | l \text{ is $\alpha$-strongly convex, $\beta$-smooth}, \abss{\Omega} = R}.
\end{align*}
In the strongly convex smooth setting, we have the following lower bound on the sum of stability and convergence rate.
\begin{theorem}
\label{theorem: lower_bound_strongly_convex}
Suppose an iterative algorithm outputs $\hat{\theta}_\iters$ at iteration $\iters$ on an empirical loss built upon a loss $l \in \lossesstronglyconvex$ and an i.i.d. sample $\samples$ of size $\obs$, and it has uniformly stability $\ErrorSta^{\hat{\theta}}(\iters, \obs, \lossesstronglyconvex)$ and has optimization error $\ErrorOpt^{\hat{\theta}}(\iters, \obs, \lossesstronglyconvex)$, then there exists a universal constant $\CCermstrong$ such that
\begin{align*}
\ErrorSta^{\hat{\theta}}(\iters, \obs, \lossesstronglyconvex) + \ErrorOpt^{\hat{\theta}}(\iters, \obs, \lossesstronglyconvex) \geq \inf_{\tilde{\theta}} \sup_{\distribution \in \distributions}\Exs_{\samples\sim \distribution^\obs}[\delta \totalrisk(\tilde{\theta})] \geq \frac{R^2\beta}{\CCermstrong\obs}.
\end{align*}
\end{theorem}
The trade-off in the strongly convex smooth setting is similar to that of convex smooth setting, except that the minimax estimation rate is of order $O(\frac{1}{\obs})$ instead of $O(\frac{1}{\sqrt{\obs}})$. Theorem~\ref{theorem: lower_bound_strongly_convex} provides the trade-off between stability and optimization convergence rate in the strongly convex setting. Note that a similar corollary like Corollary~\ref{corollary: optimization_lower_bound}.  The proof of Theorem~\ref{theorem: lower_bound_strongly_convex} is provided in Appendix~\ref{ssec:proof_of_theorem_theorem: lower_bound_strongly_convex}.

\section{Stability of first order optimization algorithms and implications for convergence lower bounds}
\label{sec: Stability of first order optimization algorithms}
This section is devoted to establishing stability bounds of popular first order optimization algorithms and showing that our main theorem can be applied to these algorithms to obtain their convergence lower bounds. In particular, Subsection~\ref{ssec: Stability in the convex smooth setting} establishes uniform stability for first order iterative methods in the convex smooth setting and Subsection~\ref{sub:consequences_for_the_convergence_lower_bound_in_convex_smooth_setting} discusses the consequence after applying Theorem~\ref{theorem: lower_bound} to various optimization algorithms. Subsection~\ref{ssec: Stability in the strongly convex smooth setting} provides uniform stability for first order iterative algorithms in the strongly convex smooth setting and Subsection~\ref{sub:consequences_for_the_convergence_lower_bound_in_the_strongly_convex_setting} discusses the consequence after applying Theorem~\ref{theorem: lower_bound_strongly_convex} to GD and NAG.

The goal of proving uniform stability for iteration $\iters$ is to bound the difference
\begin{align*}
\abss{l(\thetahat_{S, \iters}; z) - l(\thetahat_{\samples', \iters}; z)}
\end{align*}
for the sample $\samples = (z_1, \ldots , z_k, \ldots , z_\obs)$ and the perturbed one $\samples' = (z_1, \ldots , z_k', \ldots , z_\obs)$,  uniformly for every $z \in \Zspace$. $z_1, \ldots , z_k, \ldots , z_\obs$ and $z_k'$ are drawn i.i.d from a distribution $P$. Here $\thetahat_{S, \iters}$ denotes the output model of our optimization algorithm at iteration $\iters$ based on sample $S$. The optimization algorithm is applied on a pair of data samples $\samples, \samples'$ to get two sequences of successive models $\thetahat_{S, 0}, \thetahat_{S, 0}, \ldots , \thetahat_{S, \iters}$ and $\thetahat_{S', 0}, \thetahat_{S', 1}, \ldots , \thetahat_{S', T}$. For simplicity, we use $\thetahat_t$ to denote $\thetahat_{S, t}$ and $\thetahat_t'$ for $\thetahat_{S', t}$. We first bound the model estimate difference~$\vecnorm{\thetahat_t - \thetahat_t'}{2}$, then use the $L$-Lipschitz condition of $l$ to prove stability.

Recall that the empirical loss function for data sample $\samples = (z_1, \ldots, z_\obs)$ is
\begin{align*}
\totalrisk_\samples(\thetabasic) \defeq \frac{1}{\obs} \sum_{j=1}^\obs \loss(\thetabasic; z_j) = \frac{1}{\obs} \sum_{j=1}^{\obs} \lossf_j(\thetabasic).
\end{align*}
where we have replaced $\loss(\thetabasic; z_j)$ with $\lossf_j(\thetabasic)$ to improve readability. On the other hand, the empirical loss function for the perturbed sample $\samples' = (z_1, \ldots, z_k', \ldots, z_\obs)$ is
\begin{align*}
\totalrisk_{\samples'}(\thetabasic)  = \brackets{\frac{1}{\obs} \sum_{i=1, i\neq k}^{\obs} l(\theta; z_j)} + \frac{1}{\obs} l(\theta; z_k') = \brackets{\frac{1}{\obs} \sum_{i=1, i\neq k}^{\obs} f_i(\theta)} + \frac{1}{\obs} f_k' (\theta).
\end{align*}
Remark that the two empirical loss functions only differ on one term that is proportional to the inverse of sample size $\obs$.

\subsection{Stability in the convex smooth setting}
\label{ssec: Stability in the convex smooth setting}
We establish uniform stability for gradient descent, stochastic gradient descent, Nesterov accelerated gradient method and heavy ball method with fixed momentum parameter when the loss function is convex smooth.

\subsubsection{Gradient descent (GD)}
\label{sssec: Gradient descent (GD)}
The gradient descent algorithm is an iterative method for optimization, which uses the full gradient at each iteration (See book by~\cite{boyd2004convex}). Given a convex smooth objective $F$, GD starts at some initial point $\theta_0 \in \Thetaspace$, and iterates with the following recursion
\begin{align*}
    \theta_{t+1}  = \theta_t - \eta \nabla F(\theta_t),\ t = 1, 2, \cdots,
\end{align*}
where $\eta$ is the step-size. Typically, one would choose fixed $\eta \leq \frac{1}{\beta}$ to ensure convergence~\citep{boyd2004convex}.
In the empirical risk minimization setting, the objective $F$ of the optimization is either $\totalrisk_{\samples}$ or $\totalrisk_{\samples'}$.

\begin{theorem}
\label{theorem: stability_gd_convex}
Given a data distribution $P$, under the assumption that $l(\cdot, z)$ is a convex, $L$-Lipschitz and $\beta$-smooth function for every $z \in \Zspace$, the gradient method with constant step-size $\eta \leq \frac{1}{\beta}$ on the empirical risk $\totalrisk_{\samples}$ with sample size $\obs$, which outputs $\hat{\theta}_\iters$ at iteration $\iters$, has the following uniform stability bound for all $\iters \geq 1$,
\begin{align}
    \label{eq: stability_bound_gd}
    \ErrorSta^{\text{GD}}(\hat{\theta}_\iters, l, \distribution, \obs) \leq \frac{2 \eta L^2 \iters}{\obs}.
\end{align}
\end{theorem}
We remark that this stability bound does not depend on the exact form of the loss function $l$ and the exact form of the data distribution $P$.
The proof of this theorem is provided in Appendix~\ref{sec:proof_of_theorem_theorem: stability_gd_convex}. The key step of our proof is that in such a set-up, the error caused by the difference in empirical loss functions accumulates linearly as the iteration increases. We also show in Appendix~\ref{sec:proof_of_theorem_theorem: stability_gd_convex} that this stability upper bound can be achieved by a linear loss function.

\subsubsection{Nesterov accelerated gradient methods (NAG)}
\label{sssec: Nesterov accelerated gradient methods (NAG)}
The Nesterov's accelerated gradient method attains the optimal convergence rate~$O(1/\iters^2)$ in the smooth non-strongly convex setting under the deterministic first order oracle~\citep{nesterov1983method}. Given a convex smooth objective $F$, starting at some initial point $\theta_0 = w_0 \in \Thetaspace$, NAG uses the following updates,
\begin{align*}
\theta_{t+1} &= w_t - \eta \nabla F\parenth{w_t}, \\
w_{t+1} &= \parenth{1 - \gamma_t} \theta_{t+1}  + \gamma_t \theta_t,
\end{align*}
where $\eta \leq \frac{1}{\beta}$ is the step-size. The parameter $\gamma_t$ is defined by the following recursion
\begin{align*}
\lambda_0 = 0, \lambda_t = \frac{1 + \sqrt{1 + 4 \lambda_{t-1}^2}}{2}, \text{ and } \gamma_t = \frac{1-\lambda_t}{\lambda_{t+1}},
\end{align*}
satisfying $ -1 < \gamma_t \leq 0 $. We only provide a uniform stability bound for NAG when the empirical risk function is quadratic.
We conjecture that the same stability bound holds for general convex smooth functions.

\begin{theorem}
\label{theorem: stability_nag_convex}
Given a data distribution $P$, under the assumption that $l(\cdot, z)$ is a $L$-Lipschitz, $\beta$-smooth convex quadratic loss function defined on a bounded domain for every $z \in \Zspace$, Nesterov accelerated gradient method with fixed step-size $\eta \leq \frac{1}{\beta}$, which outputs $\hat{\theta}_\iters$ at iteration $\iters$, has the following uniform stability bound for all $\iters \geq 1$,
\begin{align}
    \label{eq: stability_bound_nag}
    \ErrorSta^{\text{NAG}} (\hat{\theta}_\iters, l, \distribution, \obs) \leq \frac{4 \eta L^2 \iters^2 }{\obs}.
\end{align}
\end{theorem}

The proof of the theorem is provided in Appendix~\ref{sec:proof_of_theorem_theorem: stability_nag_convex}. We also show in Appendix that this stability upper bound is achieved by a linear loss function. Note that unlike the full gradient method and stochastic gradient descent, the stability bound of Nesterov accelerate gradient method depends quadratically on the iteration $\iters$. Even though NAG can still have small stability when early stopping is used, its stability grows faster than that of GD at the same iteration.

\subsubsection{The heavy ball method with a fixed momentum}
\label{sssec: The heavy ball method with a fixed step-size}
The heavy ball method (HB), like NAG, is also a multi-step extension of the gradient descent method~\citep{polyak1964some}. Fixed step-size and fixed momentum parameter heavy ball method has the following updates. For $t \geq 1$,
\begin{align*}
    \theta_{t+1} = \theta_t - \eta \nabla F(\theta_t) + \gamma \parenth{\theta_t - \theta_{t-1}},
\end{align*}
with fixed $\gamma \in [0, 1), \eta \in \parenth{0, \frac{2(1-\gamma)}{\beta}}$. As for the NAG, we provide only a uniform stability bound for the heavy ball method when the empirical risk function is quadratic.
We conjecture that the same stability bound holds for general convex smooth functions.

\begin{theorem}
\label{theorem: stability_heavy_ball_fixed_convex}
Given a data distribution $P$, under the assumption that $l(\cdot, z)$ is a $L$-Lipschitz, $\beta$-smooth convex quadratic loss function defined on a bounded domain for every $z$, the heavy ball method with a fixed step-size $\eta \in \parenth{0, \frac{(1-\gamma)}{\beta}}$ and a fixed momentum parameter $\gamma \in [0, 1)$, which outputs $\hat{\theta}_\iters$ at iteration $\iters$, has the following uniform stability bound for all $\iters \geq 1$,
\begin{align}
    \label{eq: stability_bound_hb}
    \ErrorSta^{\text{HB, fixed}}(\hat{\theta}_\iters, l, \distribution, \obs) \leq \frac{4\eta L^2\iters}{(1-\sqrt{\gamma})\obs}.
\end{align}
\end{theorem}

The proof of this theorem is provided in Appendix~\ref{sec:proof_of_theorem_theorem: stability_heavy_ball_fixed_convex}. This theorem shows that the Heavy ball method with a fixed step-size and a fixed momentum parameter also uses multi-step gradients, it is more stable than NAG with a stability bound of order $O(\iters/\obs)$. This demonstrates that the multi-step setup does not necessarily lead to a similar or worse stability bound than that of NAG.

\subsubsection{Other methods with known stability}
\label{sssec: other_methods_with_known stability}
In this subsection, we restate the stability bounds of some other gradient methods in this subsection for completeness. The stability bounds stated in this subsection are not new, but they serve as basis of our discussion for their convergence lower bounds implied by Theorem~\ref{theorem: lower_bound} in Subsection~\ref{sub:consequences_for_the_convergence_lower_bound_in_convex_smooth_setting}.
\paragraph{Stochastic gradient descent (SGD) with fixed or varying step-size}
The stochastic gradient descent is a randomized iterative algorithm for optimization. Instead of using the full gradient information, it randomly chooses one data sample and updates the parameter estimate according to the gradient on that sample.
It starts at some initial point $\theta_0 \in \Thetaspace$, and iterates with the following recursion with $i$ chosen from the set $\{1,...,\obs\}$ uniformly at random:
\begin{align*}
    \theta_{t+1}  = \theta_t - \eta \nabla f_i(\theta_t), t=1, 2, \ldots
\end{align*}
\citet{hardt2015train} adapted the definition of uniform stability to randomized algorithms and showed that the fixed step-size $\eta \leq \frac{1}{\beta}$ stochastic gradient descent has a $\frac{2\eta L^2 \iters}{\obs}$-uniform stability bound in the convex, $L$-Lipschitz and $\beta$-smooth setting. According to Theorem 3.8 in~\citet{hardt2015train}, we have
\begin{align}
    \label{eq: stability_bound_sgd}
    \ErrorSta^{\text{SGD, fixed}}(\hat{\theta}_\iters, l, \distribution, \obs) \leq \frac{2 \eta L^2 \iters}{\obs}
\end{align}
for any convex $L$-Lipschitz and $\beta$-smooth loss function $l$.
This is a restatement of the result of~\citet{hardt2015train} in our notation.

\citet{hardt2015train} further considers stochastic gradient descent with decreasing step-sizes $\eta_t = t^{-\alpha}$ and shows that stochastic gradient descent with decreasing step-sizes has $\frac{2\eta L^2 \iters^{1-\alpha}}{\obs}$-uniform stability in the same setting.

\paragraph{Stochastic gradient Langevin dynamics (SGLD)}
\label{sssec: Stochastic Gradient Langevin Dynamics (SGLD)}

Stochastic gradient Langevin dynamics (SGLD) is a popular variant of stochastic gradient descent, where properly scaled isotropic Gaussian noise is added to an unbiased estimate of the gradient at each iteration~\citep{gelfand1991recursive}.
Stochastic gradient Langevin dynamics with temperature parameter $\temperature$ and step-size $\eta_t$, starts at some initial point $\theta_0 \in \real^\obs$, and iterates with the following recursion with $i$ chosen from the set $\braces{1,...,\obs}$ uniformly at random, and $w \sim \NORMAL(0, \Ind_\dims)$,
\begin{align*}
    \theta_{t+1} = \theta_t - \eta_t \nabla f_i (\theta_t) + \sqrt{\frac{2\eta_t}{\temperature}} w.
\end{align*}
SGLD plays an important role in sampling and optimization. It is proposed as a stochastic discrete version of the Langevin Equation $d\theta_t = - \nabla f(\theta_t) dt + \sqrt{\frac{2}{\temperature}} dB_t$, where $B_t$ is the Brownian motion.  Recent work by~\cite{raginsky2017non} has shown its effective in non-convex learning with optimization and generalization guarantees.

When SGLD is applied to optimization, a decreasing step with $\eta_t = O(\eta_0/t)$ should be used to ensure convergence to local minima. We study this particular step-size setting of SGLD.
It has been shown by~\cite{mou2017generalization} that SGLD has the following uniform stability for $L$-Lipschitz convex loss function,
\begin{align*}
    O\parenth{\frac{L}{\obs} \parenth{k_0+ L\sqrt{\temperature \sum_{t=k_0+1}^\iters \eta_t}}},
\end{align*}
where $k_0 = \min \braces{t|\eta_k \temperature L^2 < 1}$.
Plugging in the $O(\eta_0/t)$ step-size, we have that SGLD has a uniform stability bound
\begin{align}
    \label{eq: stability_bound_sgld}
    \ErrorSta^{\text{SGLD}}(\hat{\theta}_\iters, l, \distribution, \obs) \leq O(\frac{L^2 \parenth{\temperature\eta_0}^{1/2} \iters^{1/4}}{\obs}),
\end{align}
at iteration $\iters \geq 1$, for any convex $L$-Lipschitz and $\beta$-smooth loss function $l$. This is an adaptation of the result of~\citet{mou2017generalization} in our notation.

\subsection{Consequences for the convergence lower bound in convex smooth setting} 
\label{sub:consequences_for_the_convergence_lower_bound_in_convex_smooth_setting}
In this section, we apply Theorem~\ref{theorem: lower_bound} and Corollary~\ref{corollary: optimization_lower_bound} to obtain convergence lower bounds for a variety of first order optimization algorithms mentioned above. Furthermore, we compare the convergence lower bound we obtain with the known convergence upper bound for each of the optimization methods mentioned in the previous section. The known convergence upper bounds mentioned in this section can be found in the optimization textbooks (See~\citet{boyd2004convex} or~\citet{bubeck2015convex}). We also discuss how our lower bounds compare to those obtained from classical oracle model of complexity by~\cite{nemirovsky1982problem}.

Note that the assumptions in Theorem~\ref{theorem: lower_bound} are slightly different to what we use when we establish stability bounds in the previous section: the former assume bounded domain $R$ while the latter assume $L$-Lipschitz. To make these two assumptions compatible, in this subsection, we assume that the domain $R = \abss{\Thetaspace}$ is fixed and for all $z \in \Zspace$, there exists $\thetastar \in \Thetaspace$ such that $\nabla \loss(\cdot, z) = 0$. Then we have the loss is $L$-Lipschitz with $L \leq R \beta$. This is because for any $\theta \in \Thetaspace$,
\begin{align*}
    \vecnorm{\nabla l\parenth{\theta, z}}{2} =  \vecnorm{\nabla l\parenth{\theta, z} - \nabla l(\thetastar, z)}{2} \leq \beta \vecnorm{\theta - \thetastar}{2} \leq R\beta.
\end{align*}

In Table~\ref{tab: stability_convex}, we summarize all the uniform stability results and the corresponding convergence lower bound under convex smooth setting. While exact constants are provided in the main text, we only show the dependency on iteration number $\iters$ and sample size $\obs$ in the table.
\begin{table*}[h]
    \centering
    \hspace*{-1.0cm}
    {\renewcommand{\arraystretch}{1.8}
    \begin{tabular}{cccc}
        \toprule
        {\bf Method} & {\bf Uniform stability} & {\bf \shortstack{Convergence \\ upper bound (known)}} & {\bf  \shortstack{Convergence  \\ lower bound (ours)}}\\
        \midrule
        GD, $\eta = 1/\beta$ & $\displaystyle O\parenth{\frac{\iters}{\obs}}$ & $\displaystyle O\parenth{\frac{1}{\iters}}$ & $\displaystyle O\parenth{\frac{1}{\iters}}$ \\
        NAG* & $\displaystyle O\parenth{\frac{\iters^2}{\obs}}$ & $\displaystyle O\parenth{\frac{1}{\iters^2}}$ & $\displaystyle O\parenth{\frac{1}{\iters^2}}$ \\
        HB*, fixed momentum & $\displaystyle O\parenth{\frac{\iters}{\obs}}$ & $\displaystyle O\parenth{\frac{1}{\iters}}$ & $\displaystyle O\parenth{\frac{1}{\iters}}$ \\
        SGD, $\eta = 1/\beta$ & $\displaystyle O\parenth{\frac{\iters}{\obs}}$ &  $\displaystyle O\parenth{\frac{1}{\iters} + \CC} $ & $\displaystyle O\parenth{\frac{1}{\iters}}$ \\
        SGD, $\eta = O\parenth{\iters^{-\alpha}}$ & $\displaystyle O\parenth{\frac{\iters^{1-\alpha}}{\obs}}$ & $\displaystyle O\parenth{\frac{1}{\iters^{1-\alpha}}}$ & $\displaystyle O\parenth{\frac{1}{\iters^{1-\alpha}}}$ \\
        SGLD, $\eta = O\parenth{\iters^{-1}}$  & $\displaystyle O\parenth{\frac{\iters^{1/4}}{\obs}}$ &  $\displaystyle -$ & $\displaystyle O\parenth{\frac{1}{\iters^{1/4}}}$ \\
        \bottomrule
    \end{tabular}
    }
    \caption{Uniform stability and convergence lower bound under convex smooth setting. *Stability results for NAG and HB are only proved for quadratic loss and so the convergence lower bound in the same row is conjectured. $C$ is some universal constant, meaning that SGD with constant step-size does not converge to optimum. We are not aware of the convergence rate upper bound of SGLD.}
    \label{tab: stability_convex}
\end{table*}

\subsubsection{Gradient descent}
According to Equation~\eqref{eq: stability_bound_gd} in Theorem~\ref{theorem: stability_gd_convex}, the fixed-step-size full gradient method has $\frac{2\eta \parenth{R\beta}^2 \iters}{\obs}$-uniform stability. Applying Corollary~\ref{corollary: optimization_lower_bound}, knowing that its convergence does not depend on $\obs$, we obtain that its convergence rate is lower bounded by
\begin{align}
    \label{eq: opt_lower_bound_gd}
    \ErrorOpt^{\text{GD}}(\iters, \lossesconvex) \geq \frac{R^2}{2 \CCopt \eta \iters}.
    \end{align}
It is known (see e.g.~\cite{bubeck2015convex}) that for $f$ convex an $\beta$-smooth on $\real^\obs$, the full gradient method with step-size $\eta\leq\frac{1}{\beta}$ satisfies
\begin{align*}
f(\theta_\iters) - f(\theta^*) \leq \frac{2 \|\theta_0 - \theta^*\|^2} {\eta \iters}.
\end{align*}
The convergence rate lower bound obtained via our stability trade-off thus matches the known upper bound up to constant factors.

\subsubsection{Stochastic gradient descent}
According to~\cite{hardt2015train}, the fixed step-size stochastic gradient descent also has $\frac{2\eta \parenth{R \beta}^2 \iters} {\obs}$-uniform stability. Applying Corollary~\ref{corollary: optimization_lower_bound}, we obtain a convergence rate lower bound of order $O(1/\iters)$. However, it is known that fixed-step-size stochastic gradient descent can not converge arbitrarily small error at the rate $O(1/\iters)$~\citep{delyon1993accelerated}. The best rate of convergence to minimize a smooth non-strongly convex function with noisy gradients is of order $O(\iters^{-\frac{1}{2}})$~\citep{nemirovski2009robust}. Therefore, in the case of fixed step-size SGD, the convergence lower bound we provide is valid but loose. The fixed step-size SGD is a stable algorithm but is not a convergent algorithm.

On the other hand, it is shown in the same work~\citep{nemirovski2009robust} that $O(\iters^{-\frac{1}{2}})$ convergence rate is achieved by stochastic gradient descent with decreasing step-size of order $O(\iters^{-\frac{1}{2}})$. Using our stability argument, we provide insights why the stochastic gradient descent with decreasing step-size is not converging too fast. It has also been shown by~\citet{hardt2015train} that stochastic gradient descent with decreasing step-size of order $O(\iters^{-\frac{1}{2}})$ has $O(\sqrt{\iters}/\obs)$ uniform stability. Applying Corollary~\ref{corollary: optimization_lower_bound}, we conclude that when this decreasing step-size is used, gradient descent can not converge as fast as $O(\iters^{-1})$.

Similar arguments can be used to explain the conjecture by~\citet{moulines2011non} on the optimal convergence rates for stochastic gradient descent of $O(\iters^{-\alpha})$ step-size. It is shown in~\cite{moulines2011non} that, for $\alpha \in (2/3, 1)$, the convergence rate of stochastic gradient descent for the convex $\beta$-smooth case is upper bounded by $O(\iters^{\alpha - 1})$. It is shown by~\citet{hardt2015train}  that stochastic gradient descent of $O(\iters^{-\alpha})$ step-size has $O(\iters^{1-\alpha}/\obs)$ uniform stability in this set-up. Applying Corollary~\ref{corollary: optimization_lower_bound},  we provide a proof of this conjecture, confirming the optimality of this convergence rate upper bound.

\subsubsection{Nesterov accelerated gradient descent}
According to Theorem~\ref{theorem: stability_nag_convex}, the Nesterov accelerated gradient descent with fixed step-size has $\frac{4 \eta \parenth{R\beta}^2 \iters^2}{\obs}$-uniform stability for quadratic loss functions. Under the conjecture that the same stability holds for convex smooth loss functions, according to Corollary~\ref{corollary: optimization_lower_bound}, we could obtain that its convergence rate is lower bounded by
\begin{align}
    \label{eq: opt_lower_bound_nag}
    \ErrorOpt^{\text{NAG}}(\iters, \lossesconvex) \geq \frac{R^2}{4\CCopt \eta \iters^2}.
\end{align}
This is compatible with its convergence rate upper bound provided in~\cite{nesterov1983method}. For $f$ convex and $\beta$-smooth function, Nesterov accelerated gradient method with step-size $\eta\leq\frac{1}{\beta}$ satisfies
\begin{align*}
    f(\theta_\iters) - f(\theta^*) \leq \frac{2 \|\theta_1 - \theta^*\|^2}{\eta \iters^2}.
\end{align*}

We can compare our stability based lower bounds to classical ways of getting complexity lower bound using the classical first-order oracle of complexity~\citep{nemirovsky1982problem,nesterov2013introductory}. The classical oracle model based lower bound provides $O(1/\iters^2)$ lower bound for all first order optimization methods that falls into the following black-box framework. It assumes that the optimization methods takes initialization $\theta_1 = 0$ and at iteration t, $\theta_t$ is in the linear span of all previous gradients. Whereas our results show that all optimization methods with order $O(\iters^2/\obs)$ uniform stability in the smooth non-strongly convex setting would have convergence rate lower bounded by $O(1/\iters^{2})$. The two lower bounds have similar form, but apply under different scenarios. One remarkable property of our result is that it does not depend on how exactly the algorithm is initialized.

\subsubsection{Heavy ball method with fixed step-size}
According to Theorem~\ref{theorem: stability_heavy_ball_fixed_convex}, heavy ball method with fixed step-size $\eta \in \parenth{0, \frac{(1-\gamma)}{\beta}}$ and fixed momentum parameter $\gamma \in [0, 1)$ has
\begin{align*}
    \frac{4\eta L^2\iters}{(1-\sqrt{\gamma})\obs}.
\end{align*}
uniform stability for quadratic loss functions. Under the conjecture that the same stability holds for convex smooth loss functions, applying Corollary~\ref{corollary: optimization_lower_bound}, we obtain that its convergence rate is lower bounded by $O(1/\iters)$. First, this lower bound matches the convergence rate upper bound proved in~\cite{ghadimi2015global}. Second, unlike Nesterov accelerated gradient descent, even though multiple steps of gradients are used, heavy ball method with fixed step-size is not able to achieve the optimal convergence rate $O(1/\iters^2)$. Another viewpoint on this result is that the smart choice of weighting coefficients in NAG is necessary to its optimal convergence guarantee.

\subsubsection{Stochastic gradient Langevin dynamics (SGLD)}
According to~\cite{mou2017generalization}, stochastic gradient Langevin dynamics with temperature $\temperature$ and decreasing step-size $O(1/\iters)$, when used for convex optimization, has
\begin{align*}
    O(\frac{L^2 \parenth{\kappa\eta_0}^{1/2} \iters^{1/4}}{\obs})
\end{align*}
uniform-stability. Applying Corollary~\ref{corollary: optimization_lower_bound}, we conclude that its convergence rate is lower bounded by $O(1/\iters^{1/4})$. While the additional noise added in SGLD might be helpful for certain non-convex optimization settings in escaping local minima as stated in~\cite{mou2017generalization}, SGLD has a slower worst-case convergence than the GD or SGD based on our stability argument.

\subsection{Stability in the strongly convex smooth setting}
\label{ssec: Stability in the strongly convex smooth setting}
In this subsection, we establish uniform stability for gradient descent, Nesterov accelerated gradient method in the strongly convex smooth setting. In the strongly convex smooth setting, the loss function $l(\cdot, z)$ is $\alpha$ strongly-convex, $\beta$-smooth for every $z \in \Zspace$.

\subsubsection{Gradient descent (GD)}
\label{sssec: Gradient (GD) strongly convex}
The gradient descent method in the strongly convex setting has exactly the same updates as before, given a strongly convex smooth objective $F$, for $t \geq 0$,
\begin{align*}
\theta_{t+1}  = \theta_t - \eta \nabla F(\theta_t),
\end{align*}
where $\eta \leq 1/\beta$ is the step-size. While the algorithm stays the same, the strongly convex property of the loss function allows the algorithm to have a better stability.

\begin{theorem}
\label{theorem: stability_bound_gd_strongly_convex}
Given a data distribution $P$, under the assumption that $l(\cdot, z)$ is $\alpha$-strongly convex, $\beta$-smooth and $L$-Lipschitz for every $z \in \Zspace$, the full gradient method with constant step-size $\eta \leq \frac{1}{\beta}$, which outputs $\hat{\theta}_\iters$ at iteration $\iters \geq 1$, has uniform stability
\begin{align}
    \label{eq: stability_bound_gd_strongly_convex}
    \ErrorSta^{\text{GD, strongly convex}}(\hat{\theta}_\iters, l, \distribution, \obs)
    \leq \frac{4 L^2}{\alpha \obs} \parenth{1 - \parenth{1 - \frac{\eta \beta }{1+\kappa}}^\iters}.
    \end{align}
\end{theorem}
The proof of this theorem is provided in Appendix~\ref{sec:proof_of_theorem_theorem: stability_bound_gd_strongly_convex}.

\subsubsection{Stochastic gradient descent (SGD) with fixed step-size}
\label{sssec: Stochastic gradient descent (SGD) with fixed step-size}
The stochastic gradient descent in the strongly convex setting has the exactly same updates as before.
It starts at some initial point $\theta_0 \in \Thetaspace$, and iterates with the following recursion with $i$ chosen from the set $\braces{1,...,\obs}$ uniformly at random,
\begin{align*}
    \theta_{t+1}  = \theta_t - \eta \nabla f_i(\theta_t).
\end{align*}
The stability of SGD under strongly convex setting has been first discussed in~\cite{hardt2015train}. According to Theorem 3.10 in~\citet{hardt2015train}, the stability of SGD under strongly convex setting is upper bounded by
\begin{align}
    \label{eq: stability_bound_sgd_strongly_convex}
    \ErrorSta^{\text{SGD, fixed, strongly convex}}(\hat{\theta}_\iters, l, \distribution, \obs) \leq \frac{2 L^2}{\alpha \obs} \parenth{1 - \parenth{1 - \eta\alpha/2}^{\iters}}
\end{align}
at iteration $\iters \geq 1$, for any $\alpha$-strongly convex, $L$-Lipschitz and $\beta$-smooth loss function $l$.

\subsubsection{Nesterov accelerated gradient descent (NAG)}
\label{sssec: Nesterov accelerated gradient descent (NAG) strongly convex}
Unlike in the convex smooth setting, Nesterov’s accelerated gradient descent can take fixed momentum parameter in the strongly convex smooth setting.
\begin{align*}
\theta_{t+1} &= w_t - \eta \nabla F\parenth{w_t} \\
w_{t+1} &= \parenth{1+\frac{\sqrt{\kappa} - 1}{\sqrt{\kappa}+1}} \theta_{t+1} - \frac{\sqrt{\kappa} - 1}{\sqrt{\kappa} + 1} \theta_t,
\end{align*}
where $\eta \leq \frac{1}{\beta}$ is the step-size, $\kappa = \beta/\alpha$.

We prove its uniform stability for $\alpha$ strongly-convex, $\beta$-smooth for quadratic loss function.

\begin{theorem}
\label{theorem: stability_bound_nag_strongly_convex}
Given a data distribution $\distribution$, under the assumption that $l(\cdot, z)$ is $\alpha$-strongly convex, $\beta$-smooth and $L$-Lipschitz for every $z \in \Zspace$, Nesterov accelerated gradient descent method described above, which outputs $\hat{\theta}_\iters$ at iteration $\iters \geq 1$, has uniform stability
\begin{align}
    \label{eq: stability_bound_nag_strongly_convex}
    \ErrorSta^{\text{NAG, strongly convex}}(\hat{\theta}_\iters, l, \distribution, \obs) \leq \frac{4 L^2}{\alpha \obs} \parenth{1 - \parenth{1 - \frac{1}{\sqrt{\kappa}}}^{\iters}}.
\end{align}
\end{theorem}
The proof of this theorem is provided in Appendix~\ref{sec:proof_of_theorem_theorem: stability_bound_nag_strongly_convex}.

\subsection{Consequences for the convergence lower bound in the strongly convex setting} 
\label{sub:consequences_for_the_convergence_lower_bound_in_the_strongly_convex_setting}
In this subsection, we obtain convergence lower bound for GD and NAG in the $\alpha$-strongly convex $\beta$-smooth setting via Theorem~\ref{theorem: lower_bound_strongly_convex}. In Table~\ref{tab: stability_strongly_convex}, we summarize all the uniform stability results and the corresponding convergence lower bounds under strongly convex smooth setting. While exact constants are provided in the main text, we only show the dependency on iteration number $\iters$ and sample size $\obs$ in the table.
\begin{table*}[h]
    \centering
    \hspace*{-1.0cm}
    {\renewcommand{\arraystretch}{1.8}
    \begin{tabular}{cccc}
        \toprule
        {\bf Method} & {\bf Uniform stability} & {\bf \shortstack{Convergence \\ upper bound (known)}} & {\bf  \shortstack{Convergence  \\ lower bound (ours)}}\\
        \midrule
        GD & $\displaystyle O\parenth{\frac{1}{\obs} \parenth{1-e^{-O\parenth{\iters/\kappa}}}} $ & $\displaystyle e^{-O\parenth{\iters/\kappa}}$ & $\displaystyle e^{-O\parenth{\iters/\kappa}} - \CC$ \\
        NAG* & $\displaystyle O\parenth{\frac{1}{\obs} \parenth{1-e^{-O\parenth{\iters/\sqrt{\kappa}}}}} $ & $\displaystyle e^{-O\parenth{\iters/\sqrt{\kappa}}}$ & $\displaystyle e^{-O\parenth{\iters/\sqrt{\kappa}}} - C$ \\
        SGD & $\displaystyle O\parenth{\frac{1}{\obs} \parenth{1-e^{-O\parenth{\iters/\kappa}}}}$ & $\displaystyle e^{-O\parenth{\iters/\kappa}}+C$ & $\displaystyle e^{-O\parenth{\iters/\kappa}} - \CC$ \\
    \hline
    \end{tabular}
    }
\caption{Uniform stability and convergence lower bound under strongly convex  setting. *Stability results for NAG are only proved for quadratic loss and so the convergence lower bound in the same row is conjectured. $C$ is some universal constant, meaning that SGD with constant step-size does not converge to optimum and our convergence lower bound has an undesirable offset in this setting.}
\label{tab: stability_strongly_convex}
\end{table*}

\subsubsection{Gradient descent}
According to Theorem~\ref{theorem: stability_bound_gd_strongly_convex}, gradient descent with fixed step-size $\eta$ in the strongly convex smooth setting has
\begin{align*}
    \frac{4 \parenth{R\beta}^2}{\alpha \obs} \parenth{1 - \parenth{1 - \frac{\eta \beta }{1+\kappa}}^\iters}
\end{align*}
uniform stability. We apply Theorem~\ref{theorem: lower_bound_strongly_convex} to obtain a lower bound on the convergence of GD for strongly convex smooth functions.
\begin{align}
    \label{eq: opt_lower_bound_gd_strong_convex}
    \ErrorOpt^{\text{GD}}(\iters, \lossesstronglyconvex) \geq \frac{\beta R^2}{\CCermstrong \obs} - \frac{4 \parenth{R\beta}^2}{\alpha \obs} +  \frac{4 \parenth{R\beta}^2}{\alpha \obs}\parenth{1 - \frac{\eta \beta }{1+\kappa}}^{\iters}.
\end{align}

If the leading constants $\frac{\beta R^2}{\CCermstrong}$ and $\frac{4 \parenth{R\beta}^2}{\alpha}$ match, we could directly obtain a lower bound on its convergence of order $e^{-O\parenth{\iters/\parenth{1+\kappa}}}$ as we expect.
Unfortunately, due to our proof of the empirical risk minimization lower bound, a couple factors of constants are lost. Thus directly applying the stability bound makes it impossible to match the leading constants. We always have
\begin{align*}
    \frac{4 \parenth{R\beta}^2}{\alpha \obs} \geq \frac{\beta R^2}{\CCermstrong \obs}.
\end{align*}
Therefore, our trade-off result only gives convergence lower bound of GD with an offset of $\frac{\beta R^2}{\CCermstrong \obs} - \frac{4 \parenth{R\beta}^2}{\alpha \obs}$ as stated in Equation~\eqref{eq: opt_lower_bound_gd_strong_convex}.

Remark that a similar lower bound can be obtained for stochastic gradient descent using exactly the same argument for GD.

\subsubsection{Nesterov accelerated gradient descent}
According to Theorem~\ref{theorem: stability_bound_nag_strongly_convex}, Nesterov accelerated gradient descent with fixed step-size $\eta$ in the strongly convex smooth setting has
\begin{align*}
    \frac{4 L^2}{\alpha \obs} \parenth{1 - \parenth{1 - \frac{1}{\sqrt{\kappa}}}^{\iters}}
\end{align*}
uniform stability for quadratic loss function.
Since the construction of the minimax lower bound in Theorem~\ref{theorem: lower_bound_strongly_convex} is based on quadratic loss functions, applying Theorem~\ref{theorem: lower_bound_strongly_convex} by restricting to quadratic loss functions, we obtain an expected convergence lower bound of order $e^{-O\parenth{\iters/\sqrt{\kappa}}}$ with an offset,
\begin{align}
    \label{eq: opt_lower_bound_nag_strong_convex}
    \ErrorOpt^{\text{NAG}}(\iters, \lossesstronglyconvex) \geq \frac{\beta R^2}{\CCermstrong \obs} - \frac{4 \parenth{R\beta}^2}{\alpha \obs} +  \frac{4 \parenth{R\beta}^2}{\alpha \obs}\parenth{1 - \frac{1}{\sqrt{\kappa}}}^{\iters}.
\end{align}

\section{Simulations Experiments} 
\label{sec:simulations}
In this section, we first show via simulation results of a simple logistic regression applied on breast-cancer-wisconsin dataset that the stability bounds established in this paper have the right scaling on the iteration number~$\iters$. Second, we illustrate via a logistic regression problem that the stability bound characterize better the generalization error than simple uniform convergence bound at least for the first iterations of GD and NAG.

\subsection{Algorithmic Stability Rate Scaling} 
\label{sub:algorithmic_stability_rate_scaling}
We evaluate our stability bounds for all gradient methods mentioned on logistic regression with the binary classification datasets breast-cancer-wisconsin~\citep{wolberg1990multisurface}. This dataset has sample size $\obs = 699$ and dimension $\dims = 10$. The problem of logistic regression is formulated as follows.

Given a set of i.i.d. samples $\braces{(X_i, Y_i)}_{i=1}^\obs$,
with $X_i \in \real^\dims$ and $Y_i \in \braces{0, 1}$,
we want to estimate the parameter $\theta$ which
characterizes the conditional distribution of $Y_1$ given $X_1$:
\begin{align*}
    \Prob(Y_i = 1 \vert X_i; \theta) = r(\theta, X_i) = \frac{e^{\theta^\top X_i}}{1 + e^{\theta^\top X_i}}.
\end{align*}
Let $\yvec = \parenth{Y_1, \ldots, Y_\obs}^\top \in \braces{0, 1}^\obs$
and $\Xmat$ be the $\obs \times \dims $ matrix with $X_i$ as
$i^{\text{th}}$-row. The log-likelihood function we optimize over is as follows,
\begin{align}
    \label{eq:logistic_log_likelihood}
    f(\theta) = \frac{1}{\obs} \parenth{- \yvec^\top \Xmat \theta + \sum_{i=1}^\obs \log\parenth{1 + e^{\theta^\top X_i}}}.
\end{align}

\begin{figure}[h] 
\centering
\includegraphics[width=0.80\linewidth]{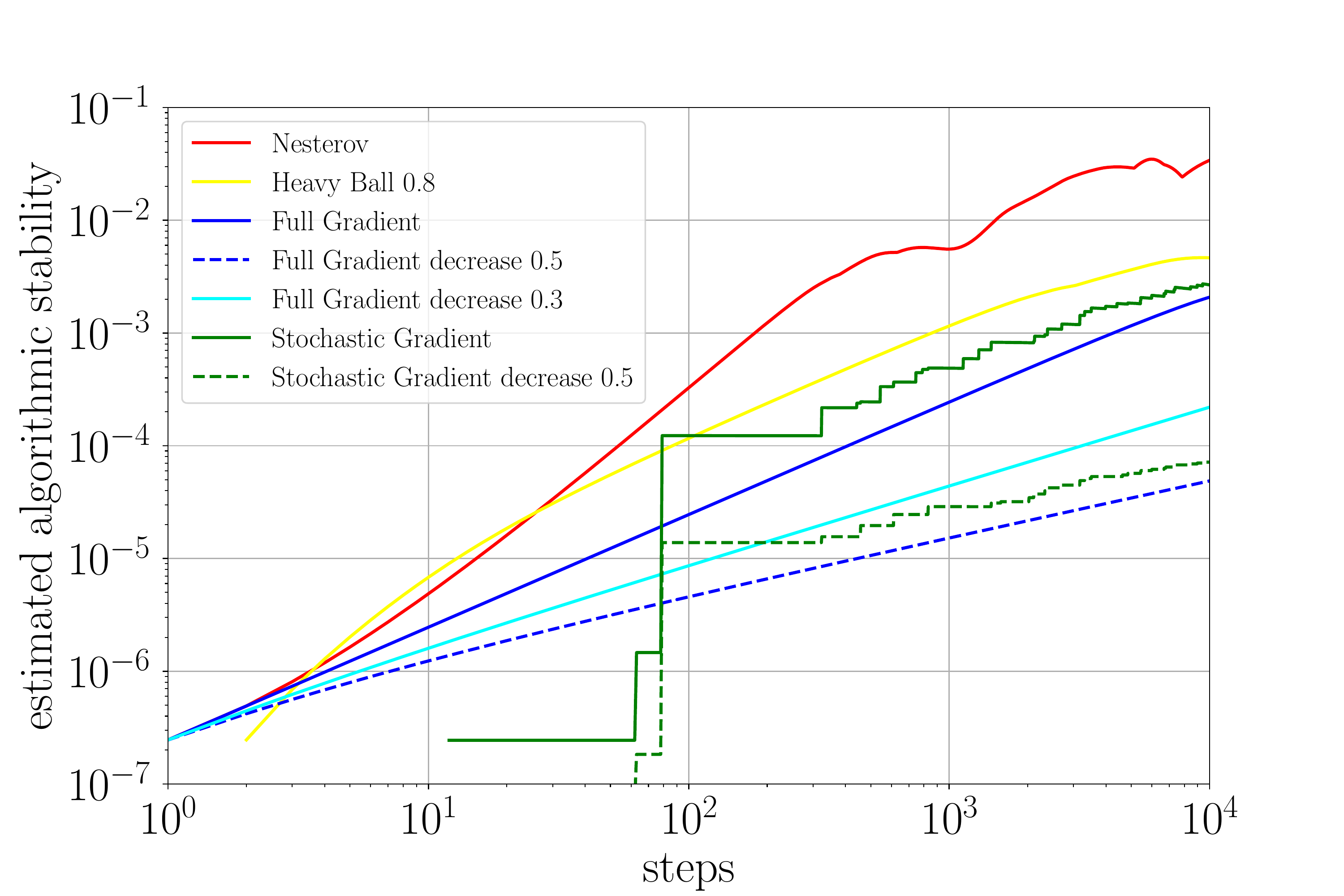}
\caption[]{Estimated algorithmic stability of various gradient methods mentioned with independent 50 runs. The estimated uniform stabilities of full gradient method, stochastic gradient method and heavy ball method with fixed step-size all have slope $1$ in log-log plot, while Nesterov accelerated gradient method has slope $2$. Methods with decreasing step-size have a slope smaller than $1$.}
\label{fig:experiment_rep}
\end{figure}

It can be shown that this objective has the Lipschitz constant $L$ equal to $1$ and the smoothness parameter $\beta$ equal to $1/4$ when the covariate matrix $\Xmat$ is normalized to have its maximum eigenvalue equal to $1$. When there is no regularization, each loss function $f_i$ is not strongly convex $\mu = 0$. In all of our experiments we set constant step-size $\eta = 0.1$. To construct samples that differ only on one data point, we first fix a sample $S$ with size $500$ from dataset, then construct a perturbed sample $S'$ by changing one data point in $S$ and finally run our optimization algorithm to compute and plot the model difference $\vecnorm{\theta_t - \theta_t'}{2}$. The norm difference $\vecnorm{\theta_t - \theta_t'}{2}$ constitute an estimate for the uniform stability up to constants independent of $\iters$ and $\obs$. Finally, the perturbation on the sample is repeated $50$ times. Figure~\ref{fig:experiment_rep} shows the estimated uniform stability, averaged over $50$ independent repeats, for all gradient methods methods, Nesterov accelerated gradient, heavy ball method with fixed momentum ($\gamma = 0.8$), full gradient method with fixed step-size, full gradient method with decreasing step-size $\iters^{-\alpha}$ ($\alpha = 0.5, 0.3$), stochastic gradient method with fixed step-size and stochastic gradient method with decreasing step-size $\iters^{-\alpha}$ ($\alpha=0.5$). We observe that the estimated uniform stabilities of full gradient method, stochastic gradient method and heavy ball method with fixed step-size all have slope $1$ in log-log plot, while Nesterov accelerated gradient method has slope $2$. As expected, methods with decreasing step-size have a slope smaller than $1$. Even though the stability bounds of NAG and HB are only established for quadratic loss, the estimated stability in the simulation makes us conjecture that the stability bounds of NAG and HB still hold in the general convex smooth setting.


\subsection{Algorithmic stability vs simple uniform convergence bounds} 
\label{sub:algorithmc_stability_vs_uniform_convergence_bounds}
The goal of this simulation is to show that algorithmic stability characterize the generalization error better than the simple uniform convergence bounds, which can not easily take into account of the growth of the function space for iterative algorithms. For $\dims$-dimensional estimation problem, simple uniform convergence bound would give an generalization error bound of order $O\parenth{\sqrt{\dims/\obs}}$. The exact constant in the uniform convergence bound depends on the function space and is hard to characterize for iterative algorithms. We think that more refined uniform convergence bound via Rademacher complexity~\citep{bartlett2003rademacher} might be possible, but we are not aware of such results for general iterative algorithms. In this section, we show via simulations that the simple uniform convergence bound of order $O\parenth{\sqrt{\dims/\obs}}$ is less precise than the stability in characterizing generalization error. More precisely, we can see that when the dimension $\dims$ and the number of samples $\obs$ are large and iteration $\iters$ is small
\begin{align*}
    \sqrt{\frac{\dims}{\obs}} \gg \frac{s(\iters)}{\obs},
\end{align*}
where $s(\iters)/\obs$ is the stability bound for GD or NAG. We show in the next two experiments that this comparison is valid and the stability bound is more relevant in large scale problems.

\begin{figure}[t]
  \centering
    \begin{minipage}{0.49\textwidth}
      \centering
    \includegraphics[width=1.0\textwidth]{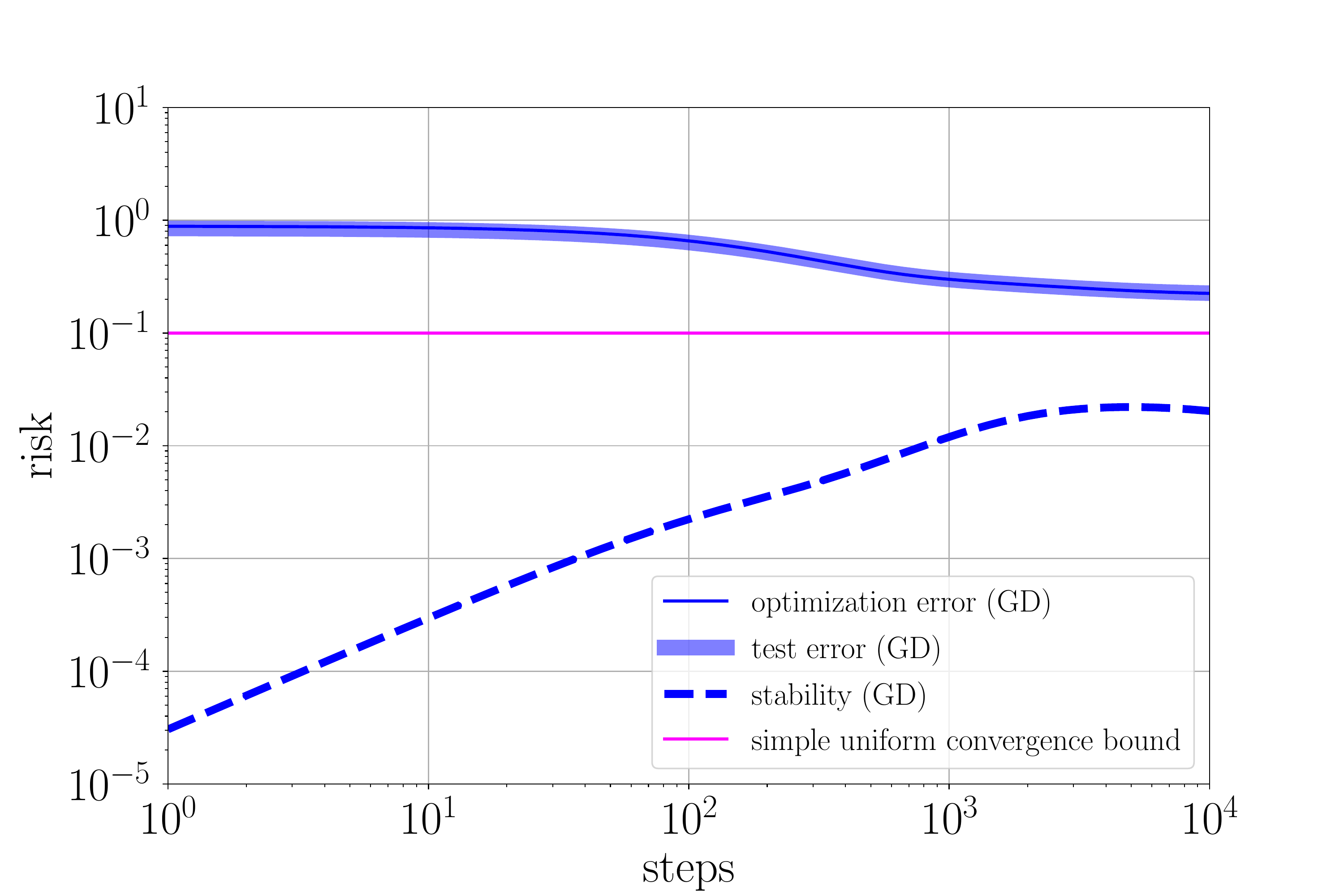}
    \subcaption[d]{Exp 1. Gradient Descent}
    \label{fig:stability_vs_ub_small_gd}
    \end{minipage}%
    \begin{minipage}{0.49\textwidth}
      \centering
    \includegraphics[width=1.0\textwidth]{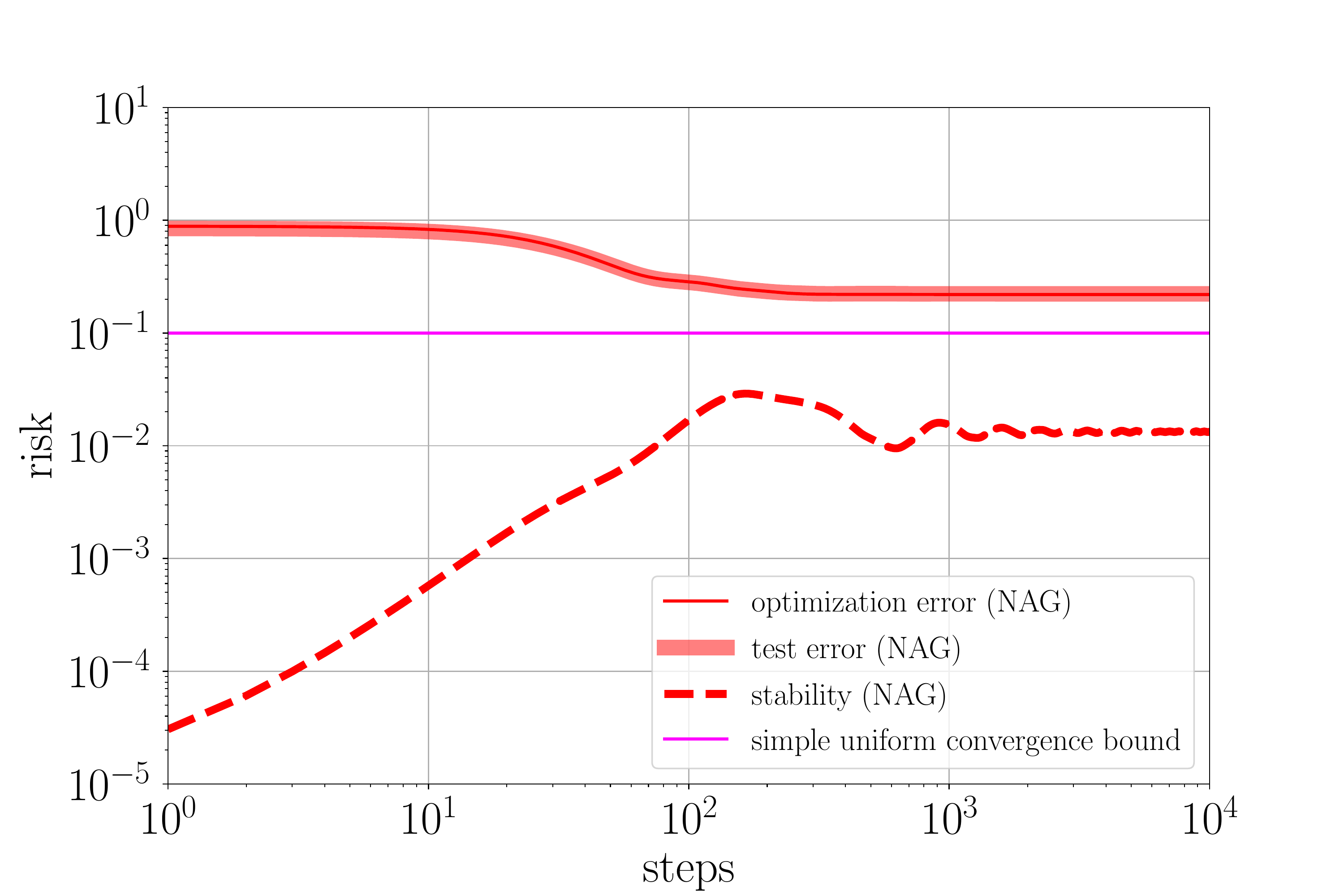}
    \subcaption[d]{Exp 1. Nesterov Accelerated Gradient}
    \label{fig:stability_vs_ub_small_nag}
    \end{minipage}
  \caption{Algorithmic stability vs simple uniform convergence bound in the first experiment, $\dims = 20, \obs = 2000$. For both GD and NAG, the optimization error plot aligns with the test error plot, indicating that optimization error dominates in the risk decomposition. Whether using stability or simple uniform convergence bound to characterize generalization error is not important. }
  \label{fig:stability_vs_ub_small}
\end{figure}

In the both experiments, we fix the true parameter $\theta^* = \parenth{1, \ldots, 1}^\top$ and we random draw $\obs$ i.i.d. samples $\parenth{X_i, Y_i}$ according to the following data generation process. Each row of $\Xmat$ is drawn from a standard $\dims$-dimensional normal distribution, and then $\Xmat$ is renormalized to have row norm $1$. Each label $Y_i$, give $X_i = x$, is drawn from a Bernoulli distribution with parameter $r(\theta^*, x)$. We use both the gradient descent and Nesterov accelerated gradient to optimize the empirical log-likelihood objective in Equation~$\eqref{eq:logistic_log_likelihood}$. We estimate the stability using its definition in Equation~\eqref{eq: stability_def} by varying different $z$ from holdout data set. In first experiment, we set $\dims = 20, \obs = 2000$. Figure~\ref{fig:stability_vs_ub_small} shows that both the simple uniform convergence bound and estimated stability bound are small compared to optimization error. In this setting, driving optimization error to zero is more important for reducing the test error, as shown in thick red color. We can still observe that the scalings of the estimated stability bound for GD and NAG are different. Our theoretical stability bound follows the estimated stability bound with the same slope, but without the saturation at the end of iterates.

\begin{figure}[t]
  \centering
    \begin{minipage}{0.49\textwidth}
      \centering
    \includegraphics[width=1.0\textwidth]{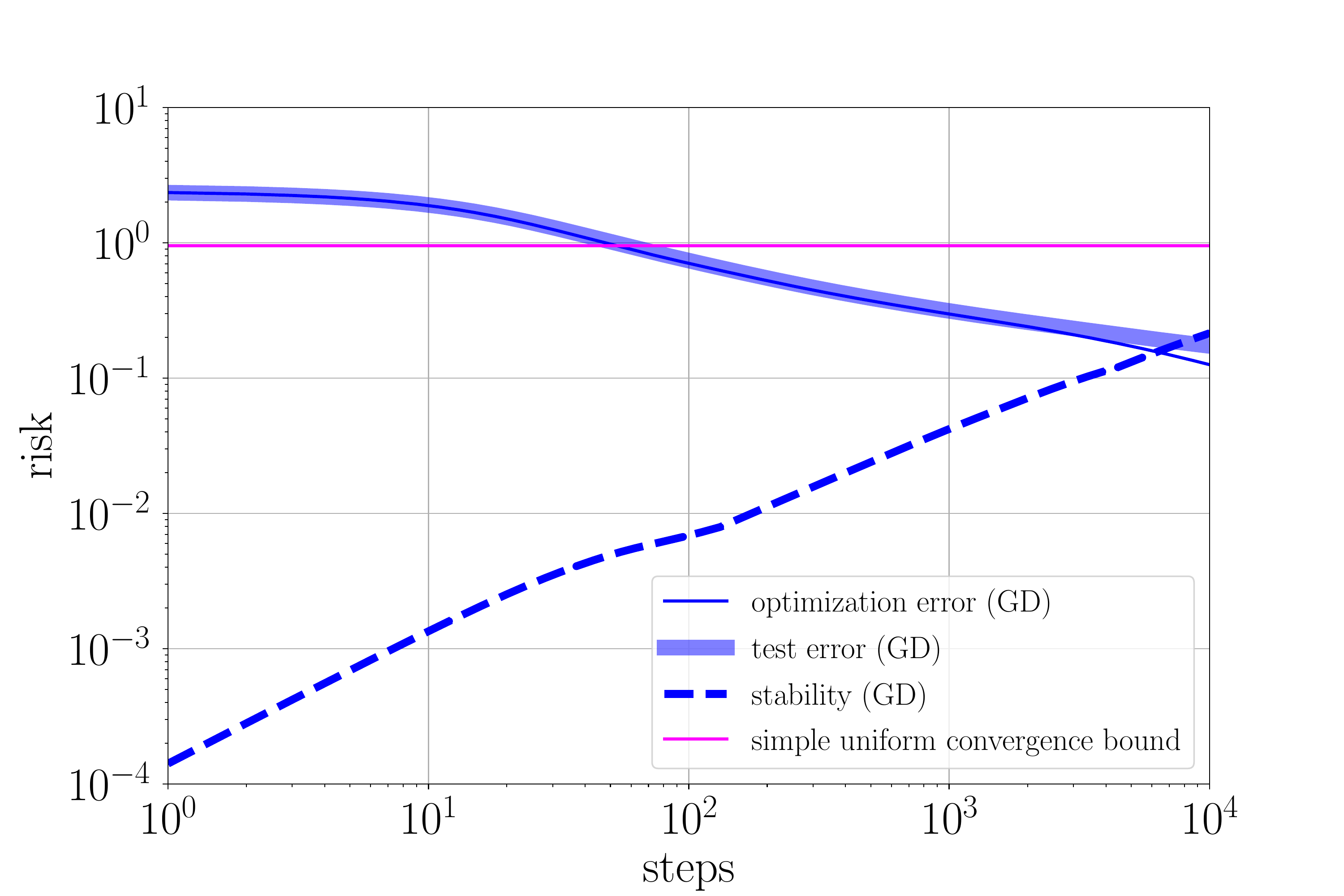}
    \subcaption[d]{Exp 2. Gradient Descent}
    \label{fig:stability_vs_ub_large_gd}
    \end{minipage}%
    \begin{minipage}{0.49\textwidth}
      \centering
    \includegraphics[width=1.0\textwidth]{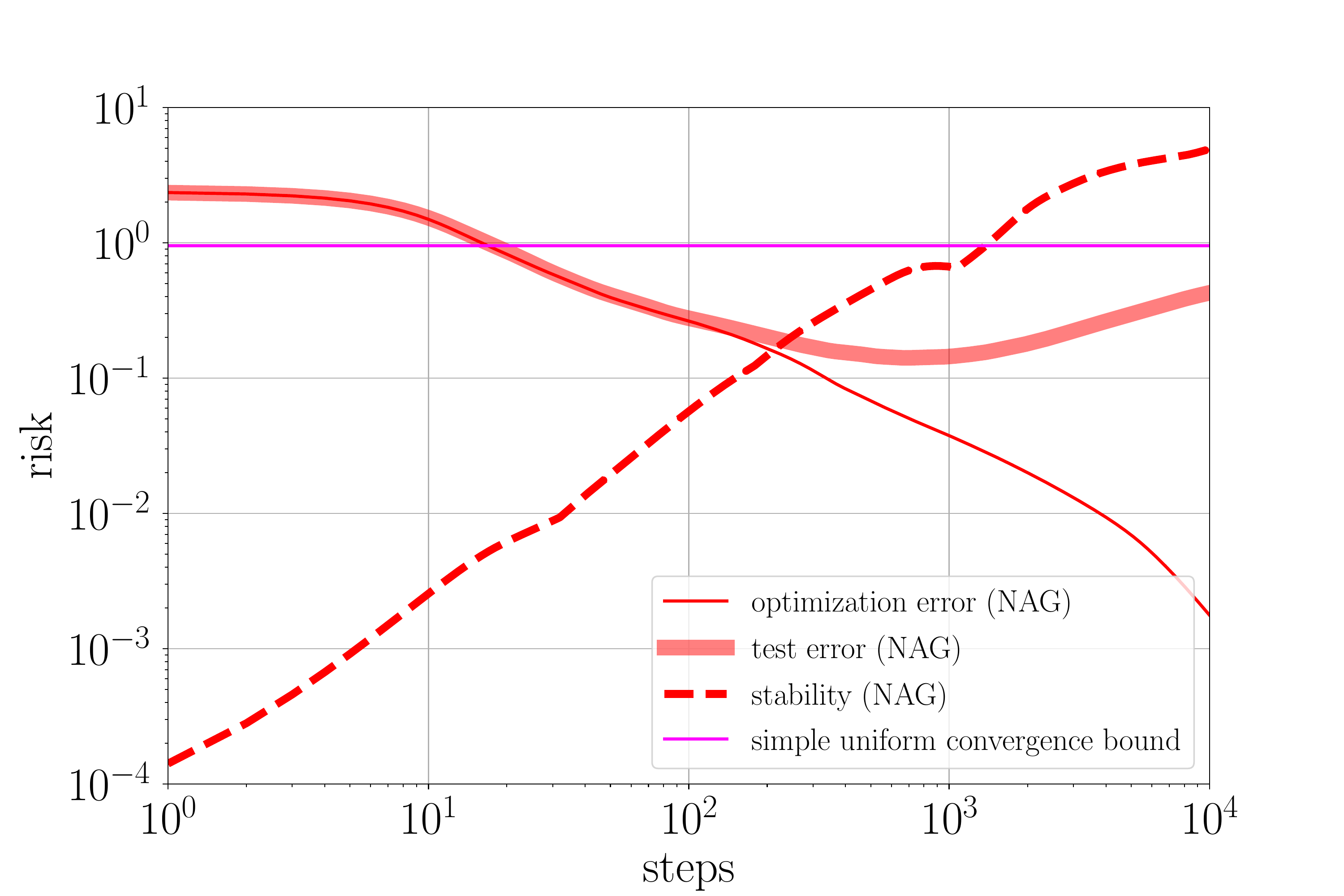}
    \subcaption[d]{Exp 2. Nesterov Accelerated Gradient}
    \label{fig:stability_vs_ub_large_nag}
    \end{minipage}
  \caption{Algorithmic stability vs simple uniform convergence bound in the second experiment, $\dims = 200, \obs = 2000$. As the test error deviates from the optimization error, the generalization error accounts for a large portion of the test error. Because the simple uniform convergence bound does not depend on the iteration number, it can't explain the overfitting phenomenon especially for NAG. }
  \label{fig:stability_vs_ub_large}
\end{figure}

\begin{figure}[t]
  \centering
    \begin{minipage}{0.49\textwidth}
      \centering
    \includegraphics[width=1.0\textwidth]{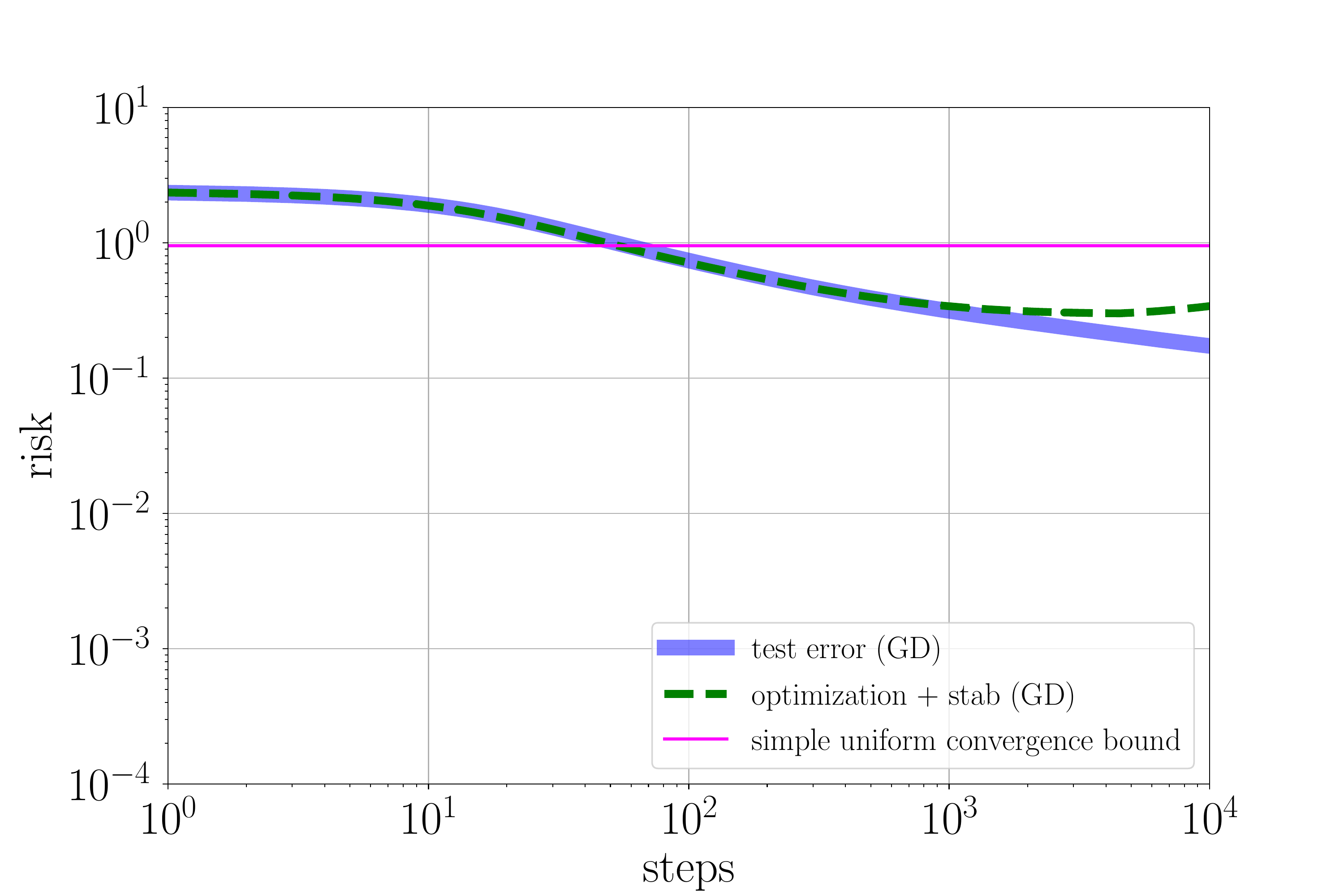}
    \subcaption[d]{Exp 2. Gradient Descent}
    \label{fig:stability_opt_large_gd}
    \end{minipage}%
    \begin{minipage}{0.49\textwidth}
      \centering
    \includegraphics[width=1.0\textwidth]{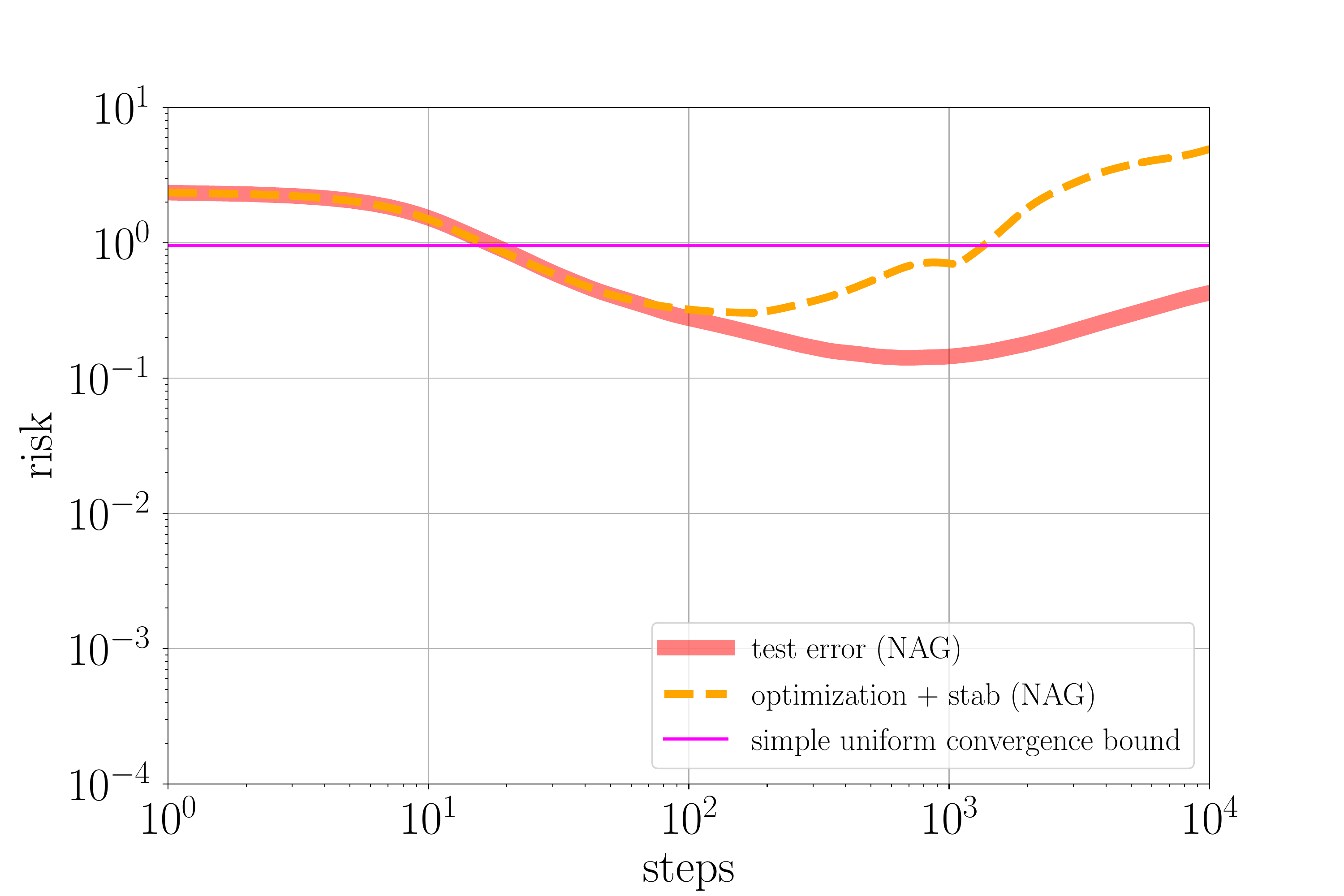}
    \subcaption[d]{Exp 2. Nesterov Accelerated Gradient}
    \label{fig:stability_opt_large_nag}
    \end{minipage}
  \caption{Stability + optimization error in the second experiment, $\dims = 200, \obs = 2000$. Stability + optimization error shown in dashed line aligns well with the test error curve. }
  \label{fig:stability_opt_large}
\end{figure}

In the second experiment, we set $\dims = 200, \obs = 2000$. Figure~\ref{fig:stability_vs_ub_large} shows that the generalization error accounts for a large portion of the test error. Especially, we observe in Figure~\ref{fig:stability_vs_ub_large_nag} that the test error of NAG deviates from its training error. Simple uniform convergence bound does not explain the overfitting phenomenon here. The algorithmic stability combined with the training error suggests that early-stopping should be used for NAG in this setting as shown in Figure~\ref{fig:stability_opt_large}.

\section{Discussions}
In this section, we discuss how our stability bound for optimization could served as an early stopping criteria. We also discuss other iterative algorithms such as boosting that could fit into this stability and optimization trade-off framework.

\subsection{Stability based early stopping criteria}
Minimizing empirical risk is often computationally expensive in large scale learning problems. As it has been pointed out in~\cite{bousquet2008tradeoffs}, optimization algorithms do not need to carry out this minimization with great accuracy since the empirical risk is already an approximation to the expected risk. For example, we can stop an iterative optimization algorithm long before its convergence to reduce computational cost. How early we should stop without deteriorating too much the expected risk becomes the main question we ask in large scale learning problems. The expected excess risk decomposition has been the main theoretical guideline for this kind of early-stopping criteria. Even though in this reasoning we are studying upper bounds of generalization and optimization
errors, it is often accepted that these upper bounds give a realistic idea of
the actual convergence rates~\citep{vapnik1994measuring,bousquet2002stability,bartlett2006convexity,bousquet2008tradeoffs}.

We would like to stop our optimization algorithm as far as it reaches an optimization error close to its generalization error.
However, the uniform convergence bounds are often too pessimistic about the size of the space to search over. Instead, we use our stability based generalization bound as an estimate of the generalization error. Formally, we would choose iteration $\iters$ such that
\begin{align*}
\ErrorSta(\iters, \obs) \approx \ErrorOpt(\iters).
\end{align*}

As an example, our stability based generalization bound for fixed-step-size full gradient method in the smooth non-strongly convex setting is $\frac{2\eta L^2 \iters}{\obs}$. The first remarkable point is that this generalization error bound is dimension-free. Because it is often hard to access accurate estimates for the uniform convergence bounds based generalization error, it might be advantageous to acquire a theoretical early-stopping criterion via our stability bounds. For the full gradient method trained model, as long as the Lipschitz constant $L$ and smoothness constant $\beta$ can be estimated accurately, we are able to give an early stopping criterion such as $ \iters \approx \sqrt{\frac{\obs}{\eta^2 L^2 R^2}}$, given the estimate of $R$ is accurate.

\subsection{Other iterative optimization algorithms such as boosting}
Boosting is one of the most successful and practical iterative optimization methods. Unlike gradient method which iterates over parameters, boosting starts with a sensible estimator or classifier, the learner, and seeks its improvements iteratively on the function space. The bias-variance trade-off of L2 boosting discussed in~\cite{buhlmann2003boosting} shares similar behaviors as the trade-off we discussed in Equation~\eqref{eq: risk}. It would be interesting to characterize the stability of boosting algorithms with various kinds of weaker learners and derive precise trade-off results as we did for gradient methods.


\acks{This research is supported in part by ONR Grant N00014-16-2664
, NSF Grants DMS-1613002 and IIS 1741340, and the Center for Science of Information (CSoI), a US NSF Science and Technology Center, under grant agreement CCF-0939370. We would like to thank Raaz Dwivedi and Rebecca Barter for fruitful discussions on this topic.}


\vskip 0.2in
\bibliography{ref}

\newpage
\appendix\label{app}

\etoctocstyle{1}{Appendix}
\etocdepthtag.toc{mtappendix}
\etocsettagdepth{mtchapter}{none}
\etocsettagdepth{mtappendix}{subsection}
\tableofcontents‎‎

\section{Proof of Main Results} 
\label{sec:proof_of_main_results}

\subsection{Proof of Theorem~\ref{theorem: lower_bound}} 
\label{ssec:proof_of_theorem_theorem: lower_bound}

Using Equation~\eqref{eq: risk}, Theorem~\ref{theorem: lower_bound} directly follows from the well-known statistical lower bound for empirical risk estimation with adaptation to convex smooth loss functions. For completeness, we restate this lower bound and provide the proof below.

\begin{lemma}
\label{lemma: erm_lower_bound}
For any fixed sample size $n$, there exists a universal constant $\CCerm>0$ and $\beta$-smooth convex loss function $l$ defined on $\Zspace \times \Thetaspace$, with $R = \abss{\Thetaspace}$, such that
\begin{align*}
\inf_{\hat{\theta}} \sup_{\distribution \in \distributions}\Exs_S\brackets{\delta R(\hat{\theta})} \geq \frac{R^2 \beta}{\CCerm\sqrt{\obs}}.
\end{align*}
\end{lemma}

\paragraph{Proof of Lemma~\ref{lemma: erm_lower_bound}} 
\label{par:proof_of_lemma_lemma: erm_lower_bound}
The main idea to prove this lemma is to formulate the excess risk minimization problem as binary hypothesis testing problem and then apply Le Cam's method for lower bound.

For any fixed sample size $\obs$, define domain~$\Zspace$ be $\braces{-1, 1}$  and two probability distributions $P_1$ and $P_2$ satisfying the following two properties,
\begin{align*}
    P_1\parenth{Z = -1} = P_2\parenth{Z = 1} = \frac{1}{2} + \frac{1}{\sqrt{24\obs}}, \\
    P_1\parenth{Z = 1} = P_2\parenth{Z = -1} = \frac{1}{2} - \frac{1}{\sqrt{24\obs}}. \\
\end{align*}
We define $P_1^\obs$ to be the joint distribution where $Z_1, \ldots, Z_n$ are independent samples from $P_1$, and we defin $P_2$ accordingly.

Let $\theta_1^* \in \Thetaspace$ with all other coordinates zero but the first coordinate equals to $-\delta$, and $\theta_2^* \in \Thetaspace$ with all other coordinates zero but the first coordinate equals to $\delta$, with $0 < \delta \leq r$. $\delta$ and $r$ are a constants to be determined later.
Let $\theta[1]$ be the first coordinate of $\theta$ and let $\Phi(r)$ be the parameter such that
\begin{align*}
    \forall v \in \braces{1, 2}, \abss{\theta[1] - \theta_v^*[1]} \geq r \Rightarrow \Exs_{Z \sim P_v} \brackets{\delta R(\theta')} \geq \Phi(r).
\end{align*}
The exact form of $\Phi(r)$ will be determined after we define the loss function $l$.
We have
\begin{align}
    \label{eq: lower_bound_inequality}
    \inf_{\hat{\theta} \in \Thetaspace} \max_{v \in \braces{1, 2}} \Exs_{P_v}\brackets{\delta R(\hat{\theta})} \geq \Phi(r) \cdot \inf_{\hat{\theta} \in \Omega} \max_{v \in \braces{1, 2}} P_v^\obs \parenth{\abss{\hat{\theta}[1] - \theta_v^*[1]} \geq r}.
\end{align}
Le Cam's method reduce this estimation problem to binary hypothesis testing problem, then we have
\begin{align*}
    \inf_{\hat{\theta} \in \Thetaspace} \max_{v \in \braces{1, 2}} P_v^\obs \parenth{\abss{\hat{\theta}(Z_v^\obs)[1] - \theta_v^*[1]} \geq r} \geq \inf_\Psi \max_{v \in \braces{1, 2}} P_v^\obs \parenth{\Psi(Z_v^\obs) \neq v},
\end{align*}
where the infimum ranges over all testing functions $\Psi: \Zspace^\obs \rightarrow \braces{1, 2}$.

We have for any $\Psi: \Zspace^\obs \rightarrow \braces{1, 2}$ that the probability of error is
\begin{align*}
    \max_{v \in \braces{1, 2}} P_v^\obs \parenth{\Psi(Z_v^\obs) \neq v} = \frac{1}{2} P_1^\obs \parenth{\Psi\parenth{Z_1^\obs} \neq 1} + \frac{1}{2} P_2^\obs \parenth{\Psi\parenth{Z_2^\obs} \neq 2}
\end{align*}
A standard result of~\cite{cam1986asymptotic} gives the exact expression of the minimal possible error in the above hypothesis test. We have
\begin{align*}
    \inf_\Psi \braces{P_1^\obs \parenth{\Psi\parenth{Z_1^\obs} \neq 1} +  P_2^\obs \parenth{\Psi\parenth{Z_2^\obs} \neq 2}} = 1 - \vecnorm{P_1^\obs - P_2^\obs}{\text{TV}},
\end{align*}
where $\vecnorm{\cdot}{\text{TV}}$ denotes the total variation distance. Using Pinsker's inequality, we have
\begin{align*}
    \vecnorm{P_1^\obs - P_2^\obs}{\text{TV}}^2
    &\leq 2 \text{KL}\parenth{P_1^\obs || P_2^\obs} \\
    &= \frac{\obs}{2} \text{KL}\parenth{P_1 || P_2} \\
    &\stackrel{(i)}{=} \frac{\obs}{2} \cdot \frac{1}{\sqrt{6\obs}} \log \frac{1+\frac{1}{\sqrt{6\obs}}}{1 - \frac{1}{\sqrt{6\obs}}} \\
    &\stackrel{(ii)}{\leq} \frac{\obs}{2} \cdot \frac{3}{6\obs} \\
    &= \frac{1}{4} \\.
\end{align*}
Equality $(i)$ uses the KL divergence formula between two Bernoulli distributions. Inequality $(ii)$ uses the inequality $\delta \log \frac{1+\delta}{1-\delta} \leq 3\delta^2$ for $\delta \in \brackets{0, \frac{1}{2}}$.
Thus, we show that any test $\Psi$ mistakes one of the probability distribution for the other with probability at least $\frac{1}{4}$.
\begin{align*}
    \inf_\Psi \max_{v \in \braces{1, 2}} P_v^\obs \parenth{\Psi(Z_v^\obs) \neq v} \geq \frac{1}{4}.
\end{align*}

It remains to design a $\beta$-smooth convex loss function $l$ and determine the exact form of $\Phi$. Without loss of generality, we can assume that $\Thetaspace$ is center around $0$. We define the loss function $l(\theta; z)$ to be
\begin{align*}
    l(\theta; -1) & = \begin{cases}
\frac{\beta}{2}\parenth{\theta[1] + r}^2 & \text{~~for~} \abss{\theta[1] + r} \leq \frac{r}{2} \\
\frac{\beta r}{4} \abss{\theta[1] + r} & \text{~~otherwise},
\end{cases}\\
    l(\theta; 1) & = \begin{cases}
\frac{\beta}{2}\parenth{\theta[1] - r}^2 & \text{~~for~} \abss{\theta[1] -r} \leq \frac{r}{2} \\
\frac{\beta r}{4} \abss{\theta[1] - r} & \text{~~otherwise}.
\end{cases}
\end{align*}
It is easy to verify that the loss function is convex and $\beta$-smooth for each $z$.
Then
\begin{align*}
    \Exs_{Z\sim P_1} l\parenth{\theta; Z} = \parenth{\frac{1}{2} + \frac{1}{\sqrt{24\obs}}} l(\theta; -1) + \parenth{\frac{1}{2} - \frac{1}{\sqrt{24\obs}}} l(\theta; 1).
\end{align*}
The function $\Exs_{Z\sim P_1} l\parenth{\theta; Z}$ is differentiable along the first coordinate. Its derivative is nondecreasing and vanishes on the interval $\brackets{-r, -\frac{r}{2}}$.
Thus the minimizer $\theta_1^*[1]$ falls into the interval $\brackets{-r, -\frac{r}{2}}$.

For $\theta' \in \Thetaspace$ such that $\abss{\theta'[1] - \theta_1^*[1]} \geq r$, using the derivative of $\Exs_{Z\sim P_1} l\parenth{\theta; Z}$, we have
\begin{align*}
    \Exs_{Z \sim P_1} \brackets{\delta R(\theta')} \geq \min\braces{\Exs_{Z \sim P_1} \brackets{\delta R(0)}, \Exs_{Z \sim P_1} \brackets{\delta R(\theta_{1, \text{left}})}}
\end{align*}
where $\theta_{1, \text{left}}$ is zero everywhere but $-\frac{3r}{2}$ on the first coordinate. Then
\begin{align*}
    \Exs_{Z \sim P_1} \brackets{\delta R(\theta')} \geq \frac{\beta r^2}{\sqrt{96\obs}},
\end{align*}
and the same holds for $P_2$. Plugging $\Phi(r) = \frac{\beta r^2}{\sqrt{96\obs}}$ into Equation~\eqref{eq: lower_bound_inequality}, we can conclude that
\begin{align*}
    \inf_{\hat{\theta} \in \Thetaspace} \max_{v \in \braces{1, 2}} \Exs_{P_v}\brackets{\delta R(\hat{\theta})} \geq \frac{\beta r^2}{\sqrt{96\obs}} \cdot \frac{1}{4} \geq \frac{\beta r^2}{16 \sqrt{6 \obs}}.
\end{align*}
We remark that we can take $r$ as large as $\frac{R}{2}$. Thus we conclude that

\begin{align*}
    \inf_{\hat{\theta} \in \Thetaspace} \max_{v \in \braces{1, 2}} \Exs_{P_v}\brackets{\delta R(\hat{\theta})} \geq \frac{R^2 \beta}{256 \sqrt{6\obs}}.
\end{align*}


\subsection{Proof of Corollary~\ref{corollary: optimization_lower_bound}}
\label{ssec:proof_of_corollary: optimization_lower_bound}
Applying Theorem~\ref{theorem: lower_bound}, for any sample size $\obs$ and $\iters$, we have
\begin{align*}
\frac{s(\iters)}{\obs} + \error_{\text{optimization}}(\iters, \obs) \geq \frac{R^2 \beta}{\CCerm\sqrt{\obs}}.
\end{align*}
As we only consider optimization method designed for any convex problems, $\mathcal{E}_{\text{optimization}}$ is independent of the sample size $\obs$. This result is valid for any sample size $\obs$. We can take $\obs$ such that the following quadratic function
\begin{align*}
Q(\frac{1}{\sqrt{\obs}}) = \frac{R^2 \beta}{\CCerm\sqrt{\obs}} - \frac{s(\iters)}{\obs},
\end{align*}
is maximized to obtain the best lower bound.

Completing the square, we have
\begin{align*}
Q(\obs) = - s(\iters)\parenth{\frac{1}{\sqrt{\obs}} - \frac{R^2\beta}{2 \CCerm s(\iters)}}^2 + \frac{R^4\beta^2}{4 \CCerm^2 s(\iters)}.
\end{align*}
$\frac{2 \CCerm s(\iters)}{R^2\beta}$ would be the best choice of $\sqrt{\obs}$, but we have to ensure that $\obs$ is an integer. Since $s(\iters)$ is divergent function of $\iters$, there exists $\iters_0 \geq 1$, such that for $\iters \geq \iters_0$, we can always find integer $\obs$ satisfying
\begin{align*}
    \frac{4 \CCerm s(\iters)}{3R^2\beta} \leq \sqrt{\obs} \leq \frac{4\CCerm s(\iters)}{R^2\beta}.
\end{align*}

Plugging $\obs$, we conclude that there exists universal constant $\CCopt$, and a convex function such that for $\iters \geq \iters_0$,
\begin{align*}
\error_{\text{optimization}}(\iters, \obs) \geq \frac{R^4 \beta^2}{\CCopt  s(T)}.
\end{align*}

\subsection{Proof of Theorem~\ref{theorem: lower_bound_strongly_convex}} 
\label{ssec:proof_of_theorem_theorem: lower_bound_strongly_convex}
We prove the statistical lower bound for empirical risk estimation in the strongly convex case via similar techniques used in the proof of Lemma~\ref{lemma: erm_lower_bound}. Le Cam's argument for reducing an estimation problem to binary hypothesis testing problem is still valid. All we do is to define a $\alpha$-strongly convex $\beta$-smooth loss function $l$ and find the corresponding $\Phi(r)$. We define the loss function $l(\theta; z)$ to be
\begin{align*}
    l(\theta; -1) &= \frac{\beta}{2} \parenth{\theta[1] + r}^2, \\
    l(\theta; 1) &= \frac{\beta}{2} \parenth{\theta[1] - r}^2.
\end{align*}
$l$ is quadratic, so it is $\alpha$-strongly convex and $\beta$ smooth for each $z$. Then
\begin{align*}
    \Exs_{Z \sim P_1} l (\theta; Z) &= \parenth{\frac{1}{2} + \frac{1}{\sqrt{24\obs}}} l(\theta; -1) + \parenth{\frac{1}{2} - \frac{1}{\sqrt{24\obs}}} l(\theta; 1) \\
    & = \frac{\beta}{2} \parenth{\theta[1]^2 + \frac{2}{\sqrt{6\obs}}\theta[1] r + r^2} \\
    & = \frac{\beta}{2} \parenth{\theta[1] + \frac{r}{\sqrt{6\obs}}}^2 + \frac{\beta}{2} \parenth{r^2 - \frac{r^2}{6 \obs}}.
\end{align*}
The minimizer $\theta_1^*$ has the first coordinate equals to $-\frac{r}{\sqrt{6\obs}}$. And the minimum is $\frac{\beta}{2} \parenth{r^2 - \frac{r^2}{6 \obs}}$.

For $\theta' \in \Thetaspace$ such that $\abss{\theta'[1] - \theta_1^*[1]} \geq r$, we have
\begin{align*}
    \Exs_{Z \sim P_1} l (\theta'; Z) \geq \frac{\beta r^2}{2}.
\end{align*}
Thus, we have
\begin{align*}
    \Exs_{Z \sim P_1} \brackets{\delta R(\theta')} \geq \frac{\beta r^2 }{12\obs}
\end{align*}
The same lower bound holds for $P_2$. Plugging $\Phi(r) = \frac{\beta r^2 }{12\obs}$ into Equation~\eqref{eq: lower_bound_inequality}, we can conclude that
\begin{align*}
    \inf_{\hat{\theta} \in \Thetaspace} \max_{v \in \braces{1, 2}} \Exs_{P_v}\brackets{\delta R(\hat{\theta})} \geq \frac{\beta r^2}{12\obs} \cdot \frac{1}{4} \geq \frac{\beta r^2}{48 \obs}.
\end{align*}

We remark that we can take $r$ as large as $\frac{R}{2}$. Thus we conclude that

\begin{align*}
    \inf_{\hat{\theta} \in \Thetaspace} \max_{v \in \braces{1, 2}} \Exs_{P_v}\brackets{\delta R(\hat{\theta})} \geq \frac{R^2 \beta}{192\obs}.
\end{align*}

\section{Stability Bounds for Convex Smooth Functions}
\label{sec:appendix_stability_bounds_for_convex_smooth_functions}

In this section, we prove stability bounds of optimization algorithms (GD, NAG and heavy ball methtod) for convex smooth functions.

Before we proceed to the main proof, we state several well known lemmas about convex optimization which can be found in~\cite{boyd2004convex,bubeck2015convex}. The $\beta$-smoothness of a function directly implies the following two lemmas. These two lemmas characterize how well the gradient approximation works for $\beta$-smooth functions in terms of both upper and lower bounds.
\begin{lemma}
\label{lemma: smooth_gradient}
Let $f$ be a $\beta$-smooth function on $\Thetaspace$. Then for all $u, v \in \Thetaspace$, we have
\begin{align*}
f(u) \leq f(v) + \nabla f(v)^\top (u-v) + \frac{\beta}{2} \vecnorm{u - v}{2}^2
\end{align*}
\end{lemma}

\begin{lemma}
\label{lemma: smooth_gradient2}
Let $f$ be a convex and $\beta$-smooth function on $\Thetaspace$. Then for any $u, v \in \Thetaspace$, we have
\begin{align*}
f(u) \geq f(v) + \nabla f(v)^\top (u - v) - \frac{1}{2\beta} \vecnorm{\nabla f(u) - \nabla f(v) }{2}^2
\end{align*}
\end{lemma}

An immediate corollary could be obtained by applying from the Lemma~\ref{lemma: smooth_gradient} to $(u, v)$ and then $(v, u)$. This corollary directly implies the constracting property of the gradient decent method, which is the key component for providing its algorithmic uniform stability.
\begin{corollary}
\label{corollary: smooth_gradient}
Let $f$ be a $\beta$-smooth function on $\Thetaspace$. Then for any $u, v \in \Thetaspace$, one has
\begin{equation*}
(\nabla f(u) - \nabla f(v))^{\top} (u-v) \geq \frac{1}{\beta} \vecnorm{\nabla f(u) - \nabla f(v)}{2}^2
\end{equation*}
\end{corollary}

\subsection{Gradient Descent} 
\label{sec:proof_of_theorem_theorem: stability_gd_convex}

Recall that in order to prove the uniform stability, we need to bound the loss difference for any fixed sample $z$ at each iteration $t \geq 1$
\begin{align*}
    \abss{ l(\theta_t, z)  - l(\theta'_t, z)}.
\end{align*}
This quantity is related to the norm difference $\vecnorm{\theta_t - \theta'_t}{2}$ under the $L$-Lipschitz condition. Using the update rule of full gradient method, we obtain an recursive relation on $\vecnorm{\theta_t - \theta'_t}{2}$.
For $\eta \leq \frac{1}{\beta}$ and $t \geq 1$, we have
\begin{align}
    \label{eq: stability_gd_recursion}
    \vecnorm{\theta_t - \theta'_t}{2} &= \vecnorm{\theta_{t-1} - \eta \nabla R_S(\theta_{t-1}) - \theta'_{t-1} + \eta \nabla R_{S'}(\theta'_{t-1})}{2} \notag \\
    & \stackrel{(i)}{\leq}  \vecnorm{\theta_{t-1} - \theta'_{t-1} - \eta \frac{1}{\obs} \sum_{i=1}^{\obs} \nabla f_i (\theta_{t-1}) + \eta \frac{1}{\obs} \sum_{i=1}^{\obs} \nabla f_i (\theta'_{t-1})}{2} + \frac{\eta}{\obs} \vecnorm{ \nabla f_k(\theta'_{t-1})  - \nabla f'_k(\theta'_{t-1}) }{2} \notag\\
    & \stackrel{(ii)}{\leq} \vecnorm{\theta_{t-1} - \theta'_{t-1} - \eta \frac{1}{\obs} \sum_{i=1}^{\obs} \nabla f_i (\theta_{t-1}) + \eta \frac{1}{\obs} \sum_{i=1}^{\obs} \nabla f_i (\theta'_{t-1})}{2} + \frac{2 \eta L}{\obs} \notag\\
    & \stackrel{(iii)}{\leq} \vecnorm{\theta_{t-1} - \theta'_{t-1}}{2}+ \frac{2\eta L}{\obs}
\end{align}
The inequality $(i)$ uses triangular inequality. The inequality $(ii)$ follows from the $L$-Lipschitz condition on the perturbed gradient terms. The last inequality $(iii)$ is obtain via the contracting property of gradient descent proved in Lemma~\ref{lemma: smooth_gradient2} and its Corollary~\ref{corollary: smooth_gradient}.

Using the recursive relation, after summing Equation~\eqref{eq: stability_gd_recursion} from $1$ to $T$, we prove that the fixed-step-size full gradient method at iteration $\iters$ is $\frac{2\eta L^2 \iters}{\obs}$-uniform stable, for $\eta \leq \frac{1}{\beta}$. That is, for every $z \in \Zspace$,
\begin{align*}
\abss{ l(\theta_\iters; z) - l(\theta'_\iters; z) } \leq \frac{2\eta L^2 \iters}{\obs}.
\end{align*}
We remark that the stability of fixed-step-size full gradient method is linear as a function of iteration $T$. More generally, for gradient descent with varying step-sizes, using the same arguments, we can prove that the stability is upper bounded by the cumulative sum of all previous step-sizes at $T$.

Next, we show that this stability upper bound can be achieved by a linear function. We design the loss function $l(\theta; z)$ such that it is either $L \theta$ or $-L\theta$ depending on $z$. We define the two empirical loss functions on $\samples$ and $\samples'$,
\begin{align*}
    \totalrisk_\samples(\theta) &= \frac{1}{\obs}\sum_{j=1}^\obs L \theta = L\theta,\\
    \totalrisk_{\samples'}(\theta) &= -\frac{1}{\obs} L \theta + \frac{1}{\obs}\sum_{j=1, j\neq k}^\obs L \theta = \frac{\obs-2}{\obs} L\theta.\\
\end{align*}
The two empirical loss functions differ exactly by $\frac{2}{\obs}L\theta$.  We have for iteration $\iters$,
\begin{align*}
    \theta_\iters &= \iters \eta L + \theta_0,\\
    \theta'_\iters &= \frac{\obs-2}{\obs}\iters \eta L + \theta_0\\
\end{align*}
Then for this linear loss, for any $z\in \Zspace$,
\begin{align*}
    \abss{ l(\theta_\iters; z) - l(\theta'_\iters; z) } = \frac{2\eta L^2 \iters}{\obs}.
\end{align*}
The stability upper bound is thus tight.

\subsection{Nesterov's Accelerated Gradient Descent} 
\label{sec:proof_of_theorem_theorem: stability_nag_convex}
Recall that the Nesterov's accelerated gradient method has the following updates for $t \geq 1$:
\begin{align}
    \label{eq: nag_update}
    \theta_{t+1} = \parenth{1 - \gamma_{t-1}}\theta_t + \gamma_{t-1} \theta_{t-1} - \eta \nabla\totalrisk_\samples(\parenth{1 - \gamma_{t-1}}\theta_t + \gamma_{t-1} \theta_{t-1}),
\end{align}
where $\eta \leq \frac{1}{\beta}$ is the step-size. $\gamma_t$ is defined by the following recursion
\begin{align*}
    \lambda_0 = 0, \lambda_t = \frac{1 + \sqrt{1 + 4 \lambda_{t-1}^2}}{2}, \text{ and } \gamma_t = \frac{1-\lambda_t}{\lambda_{t+1}},
\end{align*}
satisfying $ -1 < \gamma_t \leq 0 $.
For the updates on the perturbed samples $\samples'$, we have
\begin{align}
    \label{eq: nag_update_perturbed}
    \theta'_{t+1} = \parenth{1 - \gamma_{t-1}}\theta'_t + \gamma_{t-1} \theta'_{t-1} - \eta \nabla\totalrisk_{\samples'}(\parenth{1 - \gamma_{t-1}}\theta'_t + \gamma_{t-1} \theta'_{t-1}).
\end{align}
Denote $\Delta \theta_t = \theta_t - \theta'_t$. Taking the difference of Equation~\eqref{eq: nag_update} and~\eqref{eq: nag_update_perturbed}, we have
\begin{align}
    \label{eq: nag_update_diff}
    \Delta \theta_{t+1} = \parenth{1 - \gamma_{t-1}}\Delta \theta_t + \gamma_{t-1} \Delta \theta_{t-1} - \eta \nabla^2 \totalrisk_\samples(\theta_{\text{mid},t}) \parenth{\parenth{1 - \gamma_{t-1}}\Delta \theta_t + \gamma_{t-1} \Delta \theta_{t-1}} + e_t.
\end{align}
where the error term satisfies
\begin{align*}
    e_t = \eta \nabla\totalrisk_{\samples'}(\parenth{1 - \gamma_{t-1}}\theta'_t + \gamma_{t-1} \theta'_{t-1}) - \eta \nabla\totalrisk_{\samples}(\parenth{1 - \gamma_{t-1}}\theta'_t + \gamma_{t-1} \theta'_{t-1}),
\end{align*}
and $\theta_{\text{mid},t}$ is on the path from $\parenth{1 - \gamma_{t-1}}\theta_t + \gamma_{t-1} \theta_{t-1}$ to $\parenth{1 - \gamma_{t-1}}\theta'_t + \gamma_{t-1} \theta'_{t-1}$. Note that we have used the mean value theorem to group two gradient terms.

Because $\nabla\totalrisk_{\samples'}$ and $\nabla\totalrisk_{\samples}$ only differ in one term, using the $L$-Lipschitz gradient property, we obtain an upper bound on the error term
\begin{align*}
    \vecnorm{e_t}{2} \leq \frac{2\eta L}{\obs}.
\end{align*}
In the case of quadratic objective, we can denote
\begin{align*}
    A = \eta \nabla^2 \totalrisk_\samples(\theta_{\text{mid},t}).
\end{align*}
Using the convex and $\beta$-smooth property, we have
\begin{align*}
    0 \preceq A \preceq \Ind_\dims.
\end{align*}
Then we can rewrite Equation~\ref{eq: nag_update_diff} as follows,
\begin{align*}
    \Delta \theta_{t+1} = \parenth{\Ind_\dims - A} \brackets{\parenth{1 - \gamma_{t-1}}\Delta \theta_t + \gamma_{t-1} \Delta \theta_{t-1}} + e_t.
\end{align*}
Writing this equation in matrix form, we have
\begin{align}
    \label{eq: nag_update_diff_matrix}
    \begin{pmatrix}\Delta \theta_{t+1} \\ \Delta \theta_t \end{pmatrix} =
    \begin{pmatrix}
    \parenth{1 - \gamma_{t-1}}\parenth{\Ind_\dims - A} & \gamma_{t-1}\parenth{\Ind_\dims - A} \\
    \Ind_\dims & 0 \\
    \end{pmatrix}
    \begin{pmatrix}\Delta \theta_{t} \\ \Delta \theta_{t-1} \end{pmatrix} +
    \begin{pmatrix}e_t \\ 0 \end{pmatrix}.
\end{align}
Denote $G_t = \begin{pmatrix} \parenth{1 - \gamma_{t-1}}\parenth{\Ind_\dims - A} & \gamma_{t-1}\parenth{\Ind_\dims - A} \\ \Ind_\dims & 0 \\ \end{pmatrix}$. Then we have an explicit expression of $\Delta \theta_{t+1}$ by applying the update equation~\eqref{eq: nag_update_diff_matrix} recursively, for $t\geq 1$,
\begin{align}
    \label{eq: nag_update_diff_G}
    \begin{pmatrix} \Delta \theta_{t+1} \\ \Delta \theta_t \end{pmatrix} =
    \prod_{i=1}^t G_i
    \begin{pmatrix}\Delta \theta_1 \\ \Delta \theta_0 \end{pmatrix} +
    \sum_{i=0}^{t-1} \prod_{s = t - i + 1}^t G_s
    \begin{pmatrix} e_{t-i} \\ 0 \end{pmatrix}.
\end{align}
We have used $\prod_{i=1}^t G_i$ to denote the matrix product $G_t G_{t-1} \ldots G_1$. The goal is to bound the norm of $\Delta \theta_{t+1}$. We need the following lemma on the spectral norm of $\prod_{i=1}^t G_i$ to conclude.

\begin{lemma}
\label{lemma: nag_spectral_norm_G}
Suppose $M_t =\begin{pmatrix} (1-\gamma_t)B & \gamma_t B \\ 1 & 0 \end{pmatrix}$, where $B \in \real^{\dims \times \dims}$ is a symmetric positive semi-definite matrix $0 \preceq B \preceq \Ind_\dims $ and $-1< \gamma_t < 1$. Then for all $t \geq 1$,
\begin{align*}
\matsnorm{\prod_{i=1}^t M_i}{2} \leq 2(t+1).
\end{align*}
\end{lemma}

Assuming Lemma~\ref{lemma: nag_spectral_norm_G} as given at the moment, we now complete the proof. According to Equation~\eqref{eq: nag_update_diff_G}, applying Lemma~\ref{lemma: nag_spectral_norm_G} to $G_t$, we can bound the norm of $\Delta \theta_{t+1}$,
\begin{align*}
    \vecnorm{\Delta \theta_{t+1}}{2} &\leq 2 (t+1) \frac{2 \eta L}{\obs} + \sum_{i=0}^{t-1} 2 (i+1) \frac{2 \eta L}{\obs} \\
    &= \frac{2 \eta L}{\obs} \parenth{t^2 + 3t + 1} \\
    &\leq \frac{4 \eta L}{\obs} \parenth{t+1}^2.
\end{align*}
We have used the fact that $\vecnorm{\Delta \theta_0}{2} = 0$, $\vecnorm{\Delta \theta_1}{2} \leq \frac{2 \eta L}{\obs}$ and $\vecnorm{e_t}{2} \leq \frac{2 \eta L}{\obs}$ in the first inequality. Together with the $L$-Lipschitz condition, we obtain that the Nesterov accelerated gradient method at iteration $\iters$ is
\begin{align*}
    \frac{4\eta L^2 \iters^2}{\obs}
\end{align*}
uniform stable.

Now we turn back to prove Lemma~\ref{lemma: nag_spectral_norm_G}.
\paragraph{Proof of Lemma~\ref{lemma: nag_spectral_norm_G}}
Since $B$ is symmetric positive-semidefinite, we can diagonalize $B$. There exists a common orthogonal matrix $Q$ and diagonal matrices $D$ such that
\begin{align*}
    B = Q^{-1} D Q.
\end{align*}
We have $0 \preceq D \preceq \Ind_d$.
As a consequence, $M_i$ could also be decomposed as follows,
\begin{align*}
    M_i = \begin{pmatrix} Q^{-1} & 0 \\ 0  & Q^{-1}\end{pmatrix}
    \begin{pmatrix} \parenth{1-\gamma_{i-1}} D & \gamma_{i-1} D \\ \Ind_d & 0\end{pmatrix}
    \begin{pmatrix} Q & 0 \\ 0  & Q\end{pmatrix}.
\end{align*}
Then we obtain for its product
\begin{align*}
    \prod_{i=1}^t M_i = \begin{pmatrix} Q^{-1} & 0 \\ 0  & Q^{-1}\end{pmatrix}
    \brackets{\prod_{i=1}^t \begin{pmatrix} \parenth{1-\gamma_{i-1}} D & \gamma_{i-1} D \\ \Ind_d & 0\end{pmatrix}}
    \begin{pmatrix} Q & 0 \\ 0  & Q\end{pmatrix}.
\end{align*}
We observe that $\brackets{\prod_{i=1}^t \begin{pmatrix} \parenth{1-\gamma_{i-1}} D & \gamma_{i-1} D \\ \Ind_d & 0\end{pmatrix}}$ is a block diagonal matrix. To bound the spectral norm of $\brackets{\prod_{i=1}^t \begin{pmatrix} \parenth{1-\gamma_{i-1}} D & \gamma_{i-1} D \\ \Ind_d & 0\end{pmatrix}}$, it is sufficient to bound the $2 \times 2$ matrix of the following form
\begin{align*}
    \prod_{i=1}^t H_i,
\end{align*}
where
\begin{align*}
    H_i = \begin{pmatrix} (1-\gamma_{i-1}) h & \gamma_{i-1} h \\ 1 & 0 \end{pmatrix},
\end{align*}
with $0 \leq h \leq 1$. To bound its spectral norm, we claim the following lemma.
\begin{lemma}
    \label{lemma: nag_spectral_norm_H}
    Suppose $H_i =\begin{pmatrix} (1-\gamma_{i-1}) h & \gamma_{i-1} h \\ 1 & 0 \end{pmatrix}$, where $0 \leq h \leq 1 $ and $-1<\gamma_{i-1} < 1$. Then
    \begin{align*}
    \matsnorm{\prod_{i=1}^t H_i }{2} \leq 2(t+1).
    \end{align*}
\end{lemma}
Assuming Lemma~\ref{lemma: nag_spectral_norm_H} as given at the moment, the Lemma~\ref{lemma: nag_spectral_norm_G} can be completed.
\begin{align*}
    \matsnorm{ \prod_{i=1}^t G_i}{2} \leq \matsnorm{ \brackets{\prod_{i=1}^t \begin{pmatrix} \parenth{1-\gamma_{i-1}} D & \gamma_{i-1} D \\ \Ind_d & 0\end{pmatrix}} }{2} \leq 2(t+1).
\end{align*}

Now we turn back to prove Lemma~\ref{lemma: nag_spectral_norm_H}.
\paragraph{Proof of Lemma \ref{lemma: nag_spectral_norm_H}}
Note that $\prod_{i=1}^t H_i$ is a $2 \times 2$ matrix. Let $\begin{pmatrix} a_0 \\ b_0 \end{pmatrix}$ be a vector with norm $1$. We define
\begin{align*}
    \begin{pmatrix} a_t \\ b_t \end{pmatrix} =
    \prod_{i=1}^t H_i
    \begin{pmatrix} a_0 \\ b_0 \end{pmatrix}.
\end{align*}
To bound the spectral norm of $\prod_{i=1}^t H_i$, it is sufficient to bound the norm of $\begin{pmatrix} a_t \\ b_t \end{pmatrix}$.
We going to show by recursion that
\begin{align*}
    \max\parenth{\abss{a_t}, \abss{b_t}} \leq 2(t+1).
\end{align*}

For $t=0, t=1$, the statement is easy to verify.\\
Suppose that the statement is true until $t$. We have the following recursion,
\begin{align*}
    a_{t+1} &= h\parenth{(1-\gamma_t) a_t + \gamma_t b_t} \\
    b_{t+1} &= a_t.
\end{align*}
We remark that $a_{t+1}$ as a function of $(\gamma_0, \ldots, \gamma_t)$ is a multivariate polynomial with degree one. Hence its maximum or minimum value is attained at the extreme values of the variables. Formally,
\begin{align*}
    \abss{a_{t+1}} \leq \max_{\parenth{\gamma_i}_{0 \leq i \leq t}\in \braces{-1, 1}^{t+1}} \abss{a_{t+1}(\gamma_0, \ldots, \gamma_t)}
\end{align*}

This is a combinatorial optimization problem. But we observe that there are  only four relevant cases.
\begin{itemize}
    \item If $\gamma_{t} = 1$, then we have
    \begin{align*}
        a_{t+1} &= h b_t \\
        b_{t+1} &= a_t.
    \end{align*}
    Applying the assumption of the recursion, we obtain the desired bound for $a_{t+1}$ and $b_{t+1}$.
    \item If $\gamma_{1} = 1$, then we have
    \begin{align*}
        a_1 & = h b_0 \\
        b_1 & = a_0.
    \end{align*}
    $\begin{pmatrix} a_1 \\ b_1 \end{pmatrix}$ is a vector with norm less than 1. Consider the problem with $\begin{pmatrix} a_1 \\ b_1 \end{pmatrix}$ as initialization, we obtain the desired bound for $a_{t+1}$ and $b_{t+1}$.
    \item If there exists $i \in \braces{2, \ldots, t-1}$ such that $\gamma_{i} = 1$, then
    \begin{align*}
        H_i = \begin{pmatrix} 0 & h \\ 1 & 0\end{pmatrix},
    \end{align*}
    and
    \begin{align*}
        H_{i+1} H_i H_{i-1} = h\begin{pmatrix} \parenth{1-\gamma_{i+1} + \gamma_{i+1}(1-\gamma_{i-1})}h & \gamma_{i+1}\gamma_{i-1}h \\ 1 & 0\end{pmatrix},
    \end{align*}
    Since $-1 \leq \gamma_{i+1}\gamma_{i-1} \leq 1$, this problem is again reduced to the problem where only $t-2$ matrices are multiplied together: from $H_t$ to $H_{i+2}$, then $H_{i+1} H_i H_{i-1}$, then from $H_{i-2}$ to $H_1$. We apply the assumption of the recursion and obtain the desired bound for $a_{t+1}$.
    \item Otherwise, all $\gamma_0, ..., \gamma_t$ should take value $-1$. Then
    \begin{align*}
        H_i = \begin{pmatrix} 2h & -h \\ 1 & 0\end{pmatrix}.
    \end{align*}
    Let $\begin{pmatrix} H^{11}_t & H^{12}_t \\ H^{21}_t & H^{22}_t \end{pmatrix} = \prod_{i=1}^t H_i$,
    then we have the following recursion for its entries
    \begin{align*}
        H^{11}_{i+1} &= 2 h H^{11}_i - h H^{21}_i, \\
        H^{21}_{i+1} &= H^{11}_i, \\
        H^{12}_{i+1} &= 2 h H^{12}_i - h H^{22}_i, \\
        H^{22}_{i+1} &= H^{12}_i.
    \end{align*}
    We note that $H^{11}_{i}$ satisfies the following second-order recursion
    \begin{align*}
        H^{11}_{i+1} &= 2 h H^{11}_i - h H^{11}_{i-1},
    \end{align*}
    with $H^{11}_0 = 1$ and $H^{11}_0 = 2h$. We observe that $H^{11}_{i}$ is exactly the Chebyshev polynomial~\cite{tchebychev1853theorie,mason2002chebyshev} of second kind with parameter $U_{i}(h)$. It is known that for Chebyshev polynomial of second kind,
    \begin{align*}
        U_{i}(\cos(\theta)) = \frac{\sin((i+1)\theta)}{\sin(\theta)},
    \end{align*}
    and if $z = e^{i\theta}$,
    \begin{align*}
        \abss{U_{i}(\cos(\theta))} &= \abss{\frac{z^{i+1} - z^{-i-1}}{z - z^{-1}}} \\
        & = \abss{z^{-2i}} \abss{\sum_{j=0}^i z^{2j}} \\
        & \leq i+1.
    \end{align*}
    Thus
    \begin{align*}
        \abss{ H^{11}_{t+1}} \leq t+2.
    \end{align*}
    Similarly, we show that all entries are less than $t+2$. As a consequence,
    \begin{align*}
        \max (\abss{a_{t+1}}, \abss{b_{t+1}}) \leq 2 (t+2).
    \end{align*}
\end{itemize}
This discussion of four relevant cases concludes the recursion part, and thus the proof of Lemma~\ref{lemma: nag_spectral_norm_H}.

\subsection{Heavy Ball Method with Fixed Momentum} 
\label{sec:proof_of_theorem_theorem: stability_heavy_ball_fixed_convex}
The proof of the fixed momentum heavy ball method proceeds similarly to that of the Nesterov accelerated gradient descent.

Fixed momentum heavy ball method has the following updates.
\begin{align}
    \label{eq: hb_update}
    \theta_{t+1} = \theta_t - \eta \nabla \totalrisk_{\samples'}(\theta_t) + \gamma \parenth{\theta_t - \theta_{t-1}},
\end{align}
with fixed momentum $\gamma \in [0, 1)$, and fixed step-size $\eta \in \parenth{0, \frac{(1-\gamma)}{\beta}}$.
For the updates on the perturbed samples $S'$, we have
\begin{align}
    \label{eq: hb_update_perturbed}
    \theta_{t+1}' = \theta_t' - \eta \nabla \totalrisk_{\samples'}(\theta_t') + \gamma \parenth{\theta_t' - \theta_{t-1}'}.
\end{align}
Denote $\Delta \theta_t = \theta_t - \theta_t'$. Taking the difference of Equation~\eqref{eq: hb_update} and~\eqref{eq: hb_update_perturbed}, we have
\begin{align}
    \label{eq: hb_update_diff}
    \Delta \theta_{t+1} = (1 + \gamma) \Delta \theta_t - \gamma \Delta \theta_{t-1} - \eta \nabla^2 \totalrisk_{\samples}(\theta_{\text{mid}, t})(\Delta \theta_t)+ e_t,
\end{align}
where the error term satisfies
\begin{align*}
    e_t = \eta \nabla \totalrisk_{\samples'}(\theta_t') - \eta \nabla \totalrisk_{\samples}(\theta_t'),
\end{align*}
and $\theta_{\text{mid}, t}$ is on the path from $\theta_t$ to $\theta_t'$. Here we have used the mean value theorem to group the two gradient terms and to make appear the Hessian terms. Using the $L$-Lipschitz property, we obtain an upper bound on the error term,
\begin{align*}
    \vecnorm{e_t}{2} \leq \frac{2\eta L}{\obs}.
\end{align*}
In the case of quadratic objective, we can denote
\begin{align*}
    A = \eta \nabla^2 \totalrisk_\samples(\theta_{\text{mid}, t}).
\end{align*}
Using the convex and $\beta$-smooth property, we have
\begin{align*}
    0 \preceq A \preceq \eta \beta \Ind_d.
\end{align*}
We can rewrite Equation~\eqref{eq: hb_update_diff} in matrix form,
\begin{align}
    \label{eq: hb_update_diff_matrix}
    \begin{pmatrix} \Delta \theta_{t+1} \\ \Delta \theta_t \end{pmatrix} =
    \begin{pmatrix} \parenth{1 + \gamma} \Ind - A & - \gamma \Ind \\
    \Ind & 0 \end{pmatrix}
    \begin{pmatrix} \Delta \theta_t \\ \Delta \theta_{t-1} \end{pmatrix} +
    \begin{pmatrix} e_t \\ 0 \end{pmatrix}
\end{align}
Denote $G = \begin{pmatrix} (1+\gamma)\Ind_d - A & -\gamma \Ind_d \\ \Ind_d & 0\end{pmatrix}$. Then we could obtain an explicit expression for the difference term as follows,
\begin{align}
    \label{eq: hb_update_diff_G}
    \begin{pmatrix} \Delta \theta_{t+1} \\ \Delta \theta_t \end{pmatrix} =
    \prod_{i=1}^t G_i
    \begin{pmatrix}\Delta \theta_1 \\ \Delta \theta_0 \end{pmatrix} +
    \sum_{i=0}^{t-1} \prod_{s = t - i + 1}^t G_s
    \begin{pmatrix} e_{t-i} \\ 0 \end{pmatrix}.
\end{align}
As in the proof of NAG in Appendix~\ref{sec:proof_of_theorem_theorem: stability_nag_convex}, we are going to bound the spectral norm of $\prod_{i=1}^t G_i$ to conclude. Using diagonalization of the matrices $A$, it is sufficient to consider products of the $2 \times 2$ matrices $H = \begin{pmatrix} 1 + \gamma - a & -\gamma \\ 1 & 0\end{pmatrix}$, with $0 \leq a \leq \eta\beta$. The following lemma characterizes the spectral norm of $\prod_{i=1}^t H$.
\begin{lemma}
    \label{lemma: hb_spectral_norm_H}
    Suppose $H = \begin{pmatrix} 1 + \gamma - a & -\gamma \\ 1 & 0\end{pmatrix}$, where $0< \gamma < 1$ and $0 \leq a \leq 1-\gamma $. Then
    \begin{align*}
    \matsnorm{\prod_{i=1}^t H}{2} \leq \frac{2}{1-\sqrt{\gamma}}.
    \end{align*}
\end{lemma}
Assuming Lemma~\ref{lemma: hb_spectral_norm_H} as given at the moment, we have
\begin{align*}
    \matsnorm{\prod_{i=1}^t G_i}{2} \leq \frac{2}{1-\sqrt{\gamma}}.
\end{align*}
We can complete the proof of Theorem~\ref{theorem: stability_heavy_ball_fixed_convex}.
\begin{align*}
    \vecnorm{\Delta \theta_{t+1}}{2} &\leq \frac{2}{1-\sqrt{\gamma}} \frac{2 \eta L}{\obs} + \sum_{i=0}^{t-1} \frac{2}{1-\sqrt{\gamma}} \frac{2 \eta L}{\obs} \\
    &= \frac{4 \eta L}{(1-\sqrt{\gamma})\obs} \parenth{t+1}.
\end{align*}
We have used the fact that $\vecnorm{\Delta \theta_0}{2} = 0$, $\vecnorm{\Delta \theta_1}{2} \leq \frac{2 \eta L}{\obs}$ and $\vecnorm{e_t}{2} \leq \frac{2 \eta L}{\obs}$ in the first inequality. Together with the $L$-Lipschitz condition, we obtain that the heavy ball method with fixed momentum at iteration $\iters$ is
\begin{align*}
    \frac{4\eta L^2 \iters}{(1-\sqrt{\gamma})\obs}
\end{align*}
uniform stable.

Now we turn back to prove Lemma~\ref{lemma: hb_spectral_norm_H}.
\paragraph{Proof of Lemma~\ref{lemma: hb_spectral_norm_H}}
Let $\prod_{i=1}^t H = \begin{pmatrix} a_t & b_t \\ c_t & d_t \end{pmatrix}$. We are going to show by recursion that
\begin{align*}
\max(|a_t|, |b_t|, |c_t|, |d_t|) \leq \frac{1}{1-\sqrt{\gamma}}.
\end{align*}
For $t = 0, 1$, the statement is easy to verify. \\
Suppose that the statement is true until $t$. We have by recursion formular
\begin{align*}
a_{t+1} &= ((1 + \gamma - a) a_t - \gamma c_t) \\
c_{t+1} &= a_t \\
b_{t+1} &= ((1 + \gamma - a) b_t - \gamma d_t) \\
d_{t+1} &= b_t \\
\end{align*}
with initialization $a_1 = 1 + \gamma - a, c_1 = 1, b_1 = - \gamma, d_1 = 0$. We remark that $a_i$ satisfies the following second-order recursion, for $i \geq 1$,
\begin{align*}
    a_{i+1} = (1 + \gamma - a) a_i - \gamma a_{i-1},
\end{align*}
where $a_0 = 1, a_1 = 1 + \gamma - a$. We can also add $a_{-1} = 0$.

The characteristic equation is
\begin{align*}
x^2 - \parenth{1+\gamma - a} x + \gamma = 0.
\end{align*}

The two roots are
\begin{align*}
x_{1, 2} = \frac{1+\gamma-a \pm \sqrt{\parenth{1+\gamma-a}^2 - 4 \gamma}}{2}.
\end{align*}
We note that
\begin{align*}
    \abss{x_{1, 2}} \leq 1.
\end{align*}.
We distinguish two cases based on the two roots.
\begin{itemize}
    \item The two roots are distinct. By distinct roots theorem for second order homogeneous system, we have
    \begin{align*}
        a_{t} = l_1 x_1^{t+1} + l_2 x_2^{t+1},
    \end{align*}
    where $l_1$ and $l_2$ are constants to be determined by the initial condition.
    Solving the initial condtion, we have
    \begin{align*}
        l_1 &= \frac{1}{\sqrt{\parenth{1+\gamma-a}^2 - 4\gamma}}\\
        l_2 &= -\frac{1}{\sqrt{\parenth{1+\gamma-a}^2 - 4\gamma}}.
    \end{align*}
    Hence, we can bound $a_t$ as follows,
    \begin{align*}
        \abss{a_t} & \leq \frac{1}{\abss{\sqrt{\parenth{1+\gamma-a}^2 - 4\gamma}}} \abss{x_1 - x_2} \abss{\sum_{i=0}^{t} x_1^{t-i} x_2^i} \\
        & \leq \sum_{i=0}^{t} \abss{x_2}^i \\
        & \leq \sum_{i=0}^{t} \sqrt{\gamma}^i \\
        & \leq \frac{1}{1-\sqrt{\gamma}}.
    \end{align*}
    We have used that $\abss{x_2} \leq \sqrt{\gamma}$. When the two roots have imaginary part, it is clear that $\abss{x_2} = \sqrt{\gamma}$. On the other hand, when the two roots are real, since $\abss{x_1 x_2} = \gamma$, $\abss{x_2} \leq \abss{x_1}$, we also have $\abss{x_2} \leq \sqrt{\gamma}$.
    \item The two roots are equal. $1+\gamma-a = 2\sqrt{\gamma}$.
    \begin{align*}
        x_{1, 2} = \sqrt{\gamma} < 1
    \end{align*}
    By single root theorem for second order homogeneous system, We have
    \begin{align*}
         a_{t} = (1 + t) \sqrt{\gamma}^{t} \leq \sum_{i=0}^t \sqrt{\gamma}^t \leq \frac{1}{1-\sqrt{\gamma}}.
    \end{align*}
\end{itemize}
Overall, we have proved a bound for $a_t$,
\begin{align*}
    \abss{a_t} \leq \frac{1}{1-\sqrt{\gamma}}.
\end{align*}
We can bound $b_t, c_t$ and $d_t$ similarly because they have similar recursion formular.
\begin{align*}
    \max(\abss{a_t}, \abss{b_t}, \abss{c_t}, \abss{d_t} )  \leq \frac{1}{1-\sqrt{\gamma}}.
\end{align*}
Using the relationship between spectral norm and Frobenius norm, we have
\begin{align*}
    \matsnorm{\prod_{i=1}^t H}{2} \leq \frac{2}{1-\sqrt{\gamma}}.
\end{align*}

\section{Stability Bounds for Strongly Convex Smooth Functions}
\label{sec:appendix_stability_bounds_for_strongly_convex_smooth_functions}

\subsection{Gradient Descent} 
\label{sec:proof_of_theorem_theorem: stability_bound_gd_strongly_convex}

Recall that in order to prove the uniform stability, we need bound the loss difference for any fixed sample $z$ at each iteration $t \geq 1$
\begin{align*}
    \abss{ l(\theta_t, z)  - l(\theta'_t, z)}.
\end{align*}
This quantity is related to the norm difference $\vecnorm{\theta_t - \theta'_t}{2}$ under the $L$-Lipschitz condition. Under $\alpha$-strongly-convex case, we bound $\vecnorm{\theta_t - \theta'_t}{2}$ slightly different than that in the convex smooth case.

Using the update rule of full gradient method, we obtain an recursive relation on $\vecnorm{\theta_t - \theta'_t}{2}$.
For $\eta \leq \frac{2}{\alpha+\beta}$ and $t \geq 1$, we have
\begin{align}
    \label{eq: stability_gd_recursion_strongly}
    \vecnorm{\theta_t - \theta'_t}{2} &= \vecnorm{\theta_{t-1} - \eta \nabla R_S(\theta_{t-1}) - \theta'_{t-1} + \eta \nabla R_{S'}(\theta'_{t-1})}{2} \notag \\
    & \stackrel{(i)}{\leq}  \vecnorm{\theta_{t-1} - \theta'_{t-1} - \eta \nabla R_S (\theta_{t-1}) + \eta \nabla R_S(\theta'_{t-1})}{2} + \frac{\eta}{\obs} \vecnorm{ \nabla f_k(\theta'_{t-1})  - \nabla f'_k(\theta'_{t-1}) }{2} \notag\\
    & \stackrel{(ii)}{\leq} \vecnorm{\theta_{t-1} - \theta'_{t-1} - \eta \nabla R_S (\theta_{t-1}) + \eta \nabla R_S(\theta'_{t-1})}{2} + \frac{2 \eta L}{\obs} \notag\\
    & \stackrel{(iii)}{\leq} \parenth{1 - \frac{2\alpha\beta \eta}{\alpha+\beta}}^{1/2}\vecnorm{\theta_{t-1} - \theta'_{t-1}}{2}+ \frac{2\eta L}{\obs} \notag \\
    & \stackrel{(iv)}{\leq} \parenth{1 - \frac{\alpha\beta \eta}{\alpha+\beta}} \vecnorm{\theta_{t-1} - \theta'_{t-1}}{2}+ \frac{2\eta L}{\obs}\\
\end{align}
The inequality $(i)$ uses triangular inequality. The inequality $(ii)$ follows from the $L$-Lipschitz condition on the perturbed gradient terms. The inequality $(iii)$ is obtain via the following claim, for $f$ $\alpha$-strongly convex and $\beta$-smooth, we have
\begin{align}
    \label{eq: strongly_convex_smooth_gradient}
    \parenth{\nabla f(x) - \nabla f(y)}^\top \parenth{x - y} \geq \frac{\alpha\beta}{\alpha + \beta} \vecnorm{x - y}{2}^2 + \frac{1}{\alpha + \beta} \vecnorm{\nabla f(x) - \nabla f(y)}{2}^2.
\end{align}
This claim can be easily obtain by plugging $f(x) - \frac{\alpha}{2}\vecnorm{x}{2}^2$, which is a convex function into Corollary~\ref{corollary: smooth_gradient}.
The inequality $(iv)$ uses the fact $(1-x)^{1/2} \leq 1 - x^{1/2}$, for $0 \leq x \leq 1$.

Using the recursive relation, after summing Equation~\eqref{eq: stability_gd_recursion_strongly} from $1$ to $\iters$, we have
\begin{align*}
    \vecnorm{\theta_t - \theta'_t}{2}
    &\leq \frac{2\eta L}{\obs} \parenth{\sum_{i=0}^{\iters-1} \parenth{1 - \frac{\alpha\beta \eta}{\alpha+\beta}}^i } \\
    &= \frac{4 L}{\alpha \obs} \parenth{1 - \parenth{1 - \frac{\eta \beta }{1+\kappa}}^\iters}. \\
\end{align*}
Applying the $L$-Lipschitz condition, we have for every $z \in \Zspace$,
\begin{align*}
\abss{ l(\theta_\iters; z) - l(\theta'_\iters; z) } \leq \frac{4 L^2}{\alpha \obs} \parenth{1 - \parenth{1 - \frac{\eta \beta }{1+\kappa}}^\iters}.
\end{align*}


\subsection{Nesterov's Accelerated Gradient Descent} 
\label{sec:proof_of_theorem_theorem: stability_bound_nag_strongly_convex}
According to the discussion of Equation~\ref{eq: nag_update_diff_matrix}, in the case of quadratic loss, the Nesterov accelerated gradient descent difference term is as follows
\begin{align*}
    \begin{pmatrix}\Delta \theta_{t+1} \\ \Delta \theta_t \end{pmatrix} =
    \begin{pmatrix}
    \parenth{1 + \gamma}\parenth{\Ind_\dims - A} & -\gamma\parenth{\Ind_\dims - A} \\
    \Ind_\dims & 0 \\
    \end{pmatrix}
    \begin{pmatrix}\Delta \theta_{t} \\ \Delta \theta_{t-1} \end{pmatrix} +
    \begin{pmatrix}e_t \\ 0 \end{pmatrix},
\end{align*}
where
\begin{align*}
    \gamma = \frac{\sqrt{\kappa} - 1}{\sqrt{\kappa} + 1},
\end{align*}
$\alpha \eta \Ind_\dims \leq A \leq \beta \eta \Ind_\dims$ and $\vecnorm{e_t}{2} \leq \frac{2\eta L}{\obs}$.

Denote $G = \begin{pmatrix} (1+\gamma)\parenth{\Ind_d - A} & -\gamma \parenth{\Ind_d - A} \\ \Ind_d & 0\end{pmatrix}$. Then we could obtain an explicit expression for the difference term as follows,
\begin{align}
    \label{eq: gd_strongly_convex_update_diff_G}
    \begin{pmatrix} \Delta \theta_{t+1} \\ \Delta \theta_t \end{pmatrix} =
    \prod_{i=1}^t G_i
    \begin{pmatrix}\Delta \theta_1 \\ \Delta \theta_0 \end{pmatrix} +
    \sum_{i=0}^{t-1} \prod_{s = t - i + 1}^t G_s
    \begin{pmatrix} e_{t-i} \\ 0 \end{pmatrix}.
\end{align}
As in the proof of NAG in Appendix~\ref{sec:proof_of_theorem_theorem: stability_nag_convex}, we are going to bound the spectral norm of $\prod_{i=1}^t G_i$ to conclude. Following the proof idea used in Appendix~\ref{sec:proof_of_theorem_theorem: stability_nag_convex} and Appendix~\ref{sec:proof_of_theorem_theorem: stability_heavy_ball_fixed_convex}, using diagonalization of the matrices $A$, it is sufficient to consider products of the $2 \times 2$ matrices $H = \begin{pmatrix} (1 + \gamma)h & -\gamma h \\ 1 & 0\end{pmatrix}$, with $ 1 - \beta \eta \leq h \leq 1 - \alpha \eta$. The following lemma characterizes the spectral norm of $\prod_{i=1}^t H$.
\begin{lemma}
    \label{lemma: gd_strongly_convex_spectral_norm_H}
    Suppose $H = \begin{pmatrix} (1 + \gamma)h & -\gamma h \\ 1 & 0\end{pmatrix}$, where $\gamma = \frac{\sqrt{\kappa} - 1}{\sqrt{\kappa} + 1}$ and $1 - \beta \eta \leq h \leq 1 - \alpha \eta $. Then
    \begin{align*}
    \matsnorm{\prod_{i=1}^t H}{2} \leq 2(1+t)\parenth{\gamma (1 - \alpha \eta) }^{t/2}.
    \end{align*}
\end{lemma}

Assuming Lemma~\ref{lemma: gd_strongly_convex_spectral_norm_H} as given at the moment, we have
\begin{align*}
    \matsnorm{\prod_{i=1}^t G_i}{2} \leq 2(1+t)\parenth{\gamma (1 - \alpha \eta) }^{t/2}.
\end{align*}
We can complete the proof of Theorem~\ref{theorem: stability_bound_nag_strongly_convex}.
\begin{align*}
    \vecnorm{\Delta \theta_{t+1}}{2} &\leq \frac{2 \eta L}{\obs}\parenth{2(1+t)\parenth{\gamma (1 - \alpha \eta) }^{t/2}  + \sum_{i=0}^{t-1} 2(1+i)\parenth{\gamma (1 - \alpha \eta) }^{i/2} }\\
    &= \frac{4 \eta L}{\obs}\parenth{\sum_{i=0}^{t} (1+i)\parenth{\gamma (1 - \alpha \eta) }^{i/2} }
\end{align*}
We have used the fact that $\vecnorm{\Delta \theta_0}{2} = 0$, $\vecnorm{\Delta \theta_1}{2} \leq \frac{2 \eta L}{\obs}$ and $\vecnorm{e_t}{2} \leq \frac{2 \eta L}{\obs}$ in the first inequality. Let $p = \parenth{\gamma (1 - \alpha \eta) }^{1/2}$ and
\begin{align*}
    S = \sum_{i=0}^{t} (1+i)p^{i}.
\end{align*}
Then
\begin{align*}
    (1-p)S = \sum_{i=0}^{t} p^{i} - (t+1) p^{t+1} \leq \frac{1 - p^{t+1}}{1-p}.
\end{align*}
We also have upper and lower bounds on $p$,
\begin{align*}
    p^2 = \gamma (1 - \alpha \eta) = \frac{\sqrt{\kappa} - 1}{\sqrt{\kappa}+1} \cdot \frac{\kappa - \eta \beta }{\kappa} \leq \parenth{\frac{\sqrt{\kappa} - \sqrt{\eta \beta }}{\sqrt{\kappa}}}^2,
\end{align*}
and
\begin{align*}
    p^2 \geq \parenth{\frac{\sqrt{\kappa} - 1}{\sqrt{\kappa}}}^2.
\end{align*}
Thus
\begin{align*}
    \vecnorm{\Delta \theta_{t+1}}{2} &\leq \frac{4 \eta L}{\obs}\parenth{\sum_{i=0}^{t} (1+i)\parenth{\gamma (1 - \alpha \eta) }^{i/2} }\\
    &\leq \frac{4 \eta L}{(1-p)^2 \obs} \parenth{1 - p^{t+1}} \\
    &\leq \frac{4 L}{\alpha \obs} \parenth{1 - \parenth{1 - \frac{1}{\sqrt{\kappa}}}^{t+1}}.
\end{align*}

Together with the $L$-Lipschitz condition, we obtain that the heavy ball method with fixed momentum at iteration $\iters$ is
\begin{align*}
    \frac{4 L^2}{\alpha \obs} \parenth{1 - \parenth{1 - \frac{1}{\sqrt{\kappa}}}^{T}}
\end{align*}
uniform stable.

Now we turn back to prove Lemma~\ref{lemma: gd_strongly_convex_spectral_norm_H}.
\paragraph{Proof of Lemma~\ref{lemma: gd_strongly_convex_spectral_norm_H}}
Let $\prod_{i=1}^t H = \begin{pmatrix} a_t & b_t \\ c_t & d_t \end{pmatrix}$. We are going to show by recursion that
\begin{align*}
\max(|a_t|, |b_t|, |c_t|, |d_t|) \leq (1+t)\parenth{\gamma (1 - \alpha \eta) }^{t/2}.
\end{align*}
For $t = 0, 1$, the statement is easy to verify. \\
Suppose that the statement is true until $t$. We have by recursion formular
\begin{align*}
a_{t+1} &= ((1 + \gamma) h a_t - \gamma h c_t) \\
c_{t+1} &= a_t \\
b_{t+1} &= ((1 + \gamma) h b_t - \gamma h d_t) \\
d_{t+1} &= b_t \\
\end{align*}
with initialization $a_1 = (1+\gamma) h, b_1 = -\gamma h, c_1 = 1$ and $d_1 = 0$. We remark that $a_i$, satisfies the following second-order recursion, for $i \geq 1$,
\begin{align*}
     a_{i+1} = (1+\gamma) h a_i - \gamma h a_{i-1},
\end{align*}
where $a_0 = 1, a_1 = (1+\gamma) h$. We can also add $a_{-1} = 0$.

The characteristic equation is
\begin{align*}
    x^2 - (1 + \gamma) h x + \gamma h = 0.
\end{align*}
The two roots are
\begin{align*}
    x_{1, 2} = \frac{(1+\gamma)h \pm \sqrt{(1+\gamma)^2 h^2 - 4 \gamma h}}{2}.
\end{align*}
We verify that
\begin{align*}
    \Delta = (1+\gamma)^2 h^2 - 4 \gamma h = 4 h\parenth{ \frac{\kappa h - (\kappa-1)}{(\sqrt{\kappa} + 1)^2}} \leq 0,
\end{align*}
because $h \leq 1-\alpha \eta \leq \frac{\kappa -1 }{\kappa}$. Hence either we have equal real roots, or we have complex roots with imaginary parts.

We distinguish these two cases.
\begin{itemize}
    \item The two roots are equal. $(1+\gamma) h = 2\sqrt{\gamma h}$. Then
    \begin{align*}
        x_{1, 2} = \sqrt{\gamma h} < 1.
    \end{align*}
    By single root theorem for second order homogeneous system, we have
    \begin{align*}
        a_t = (1+t) \parenth{\gamma h}^{t/2} \leq (1+t) \parenth{\gamma (1 - \alpha \eta)}^{t/2} .
    \end{align*}
    \item The two roots are distinct.
    \begin{align*}
        \abss{x_{1, 2}} = \sqrt{\gamma h} < 1.
    \end{align*}
    By distinct roots theorem for second order homogeneous system, we have
    \begin{align*}
        a_{t} = l_1 x_1^{t+1} + l_2 x_2^{t+1},
    \end{align*}
    where $l_1$ and $l_2$ are constants to be determined by the initial condition.
    Solving the initial condtion, we have
    \begin{align*}
        l_1 &= \frac{1}{\sqrt{(1+\gamma)^2 h^2 - 4 \gamma h}}\\
        l_2 &= -\frac{1}{\sqrt{(1+\gamma)^2 h^2 - 4 \gamma h}}.
    \end{align*}
    Hence, we can bound $a_t$ as follows,
    \begin{align*}
        \abss{a_t} & \leq \frac{1}{\abss{\sqrt{(1+\gamma)^2 h^2 - 4 \gamma h}}} \abss{x_1 - x_2} \abss{\sum_{i=0}^{t} x_1^{t-i} x_2^i} \\
        & \leq \sum_{i=0}^{t} \parenth{\gamma h}^{t/2} \\
        & \leq (1+t) \parenth{\gamma(1-\alpha\eta)}^{t/2}.
    \end{align*}
\end{itemize}

We can bound $b_t, c_t$ and $d_t$ similarly because they have similar recursion formular.
\begin{align*}
    \max(\abss{a_t}, \abss{b_t}, \abss{c_t}, \abss{d_t} )  \leq (1+t)\parenth{\gamma (1 - \alpha \eta) }^{t/2}..
\end{align*}
Using the relationship between spectral norm and Frobenius norm, we have
\begin{align*}
    \matsnorm{\prod_{i=1}^t H}{2} \leq 2(1+t)\parenth{\gamma (1 - \alpha \eta) }^{t/2}.
\end{align*}

\end{document}